\documentclass[applsci,accept,moreauthors,pdftex]{Definitions/mdpinologo} 

\usepackage{color, soul}
\usepackage{xcolor}

\usepackage{url}
\newcommand{\rmk}[1]{}
\newcommand{\resubmit}[1]{{#1}}

\newcommand{\removed}[1]{}
\usepackage{subfig}
\usepackage{comment}

\usepackage{hyperref} 

\usepackage{fancyhdr} 
\usepackage{algorithm}

\usepackage{stmaryrd}

\usepackage[font=small]{caption}

\usepackage{algorithm} 
\usepackage{algorithmic}

\newcounter{descriptcounti}
\setlist[description]{%
  before={\setcounter{descriptcounti}{0}},%
    style=nextline,
  ,font=\bfseries\refstepcounter{descriptcounti}\thedescriptcounti.~}
  
\firstpage{1} 
\pubvolume{11}
\issuenum{3}
\articlenumber{975}
\pubyear{2021}
\copyrightyear{2021}
\datereceived{} 
\dateaccepted{} 
\datepublished{} 
\hreflink{https://doi.org/10.3390/app11030975} 

\Title{Intrinsically Motivated Open-Ended Multi-Task Learning Using Transfer Learning to Discover Task Hierarchy}
\TitleCitation{Intrinsically Motivated Open-Ended Multi-Task Learning Using Transfer Learning to Discover Task Hierarchy}

\Author{Nicolas Duminy $^{1,2}$, Sao Mai Nguyen $^{2,3,}$*\orcidA{}, Junshuai Zhu $^{2}$, Dominique Duhaut $^{1}$ and Jerome Kerdreux $^{2}$}

\AuthorNames{Nicolas Duminy, Sao Mai Nguyen , Junshuai Zhu, Dominique Duhaut and Jerome Kerdreux}

\AuthorCitation{Duminy, N.; Nguyen, S.M.; Zhu, J.; Duhaut, D.; Kerdreux, J.}
\address{%
$^{1}$ \quad D\'epartement Math\'ematiques Informatique Statistique, Universit\'e Bretagne Sud,
 Lab-STICC, UBL, 56321 Lorient,  
 France; nicolas.duminy@telecom-bretagne.eu (N.D.); dominique.duhaut@univ-ubs.fr (D.D.)\\
$^{2}$ \quad IMT Atlantique, Lab-STICC, UBL, 29238 Brest,  
 France; junshuai.zhu@imt-atlantique.net (J.Z.); jerome.kerdreux@imt-atlantique.fr (J.K.)\\
$^{3}$ \quad Flowers Team, U2IS, ENSTA Paris, Institut Polytechnique de Paris \& Inria, 91120 Palaiseau, France
}

\corres{Correspondence: nguyensmai@gmail.com}

\abstract{In open-ended continuous environments, robots need to learn multiple parameterised control tasks in hierarchical reinforcement learning. We hypothesise that the most complex tasks can be learned more easily by transferring knowledge from simpler tasks, and faster by adapting the complexity of the actions to the task.  We propose a task-oriented representation of complex actions, called \textit{procedures}, 
 to learn online task relationships and unbounded sequences of action primitives to control the different observables of the environment. 
Combining both goal-babbling with imitation learning, and active learning with transfer of knowledge based on intrinsic motivation,  our algorithm self-organises its learning process. It chooses at any given time a task to focus on; and what, how, when and from whom to transfer knowledge.   
We show with a simulation and a real industrial robot arm, in cross-task and cross-learner transfer settings, that task composition is key to tackle highly complex tasks.  Task decomposition is also efficiently transferred across different embodied learners and by active imitation, where the robot requests just a small amount of demonstrations and the adequate type of information. The robot learns and exploits task dependencies so as to learn tasks of every complexity.
 }


\begin{document}

\section{Introduction}

\begin{figure}[H]
\includegraphics[width=0.7\linewidth]{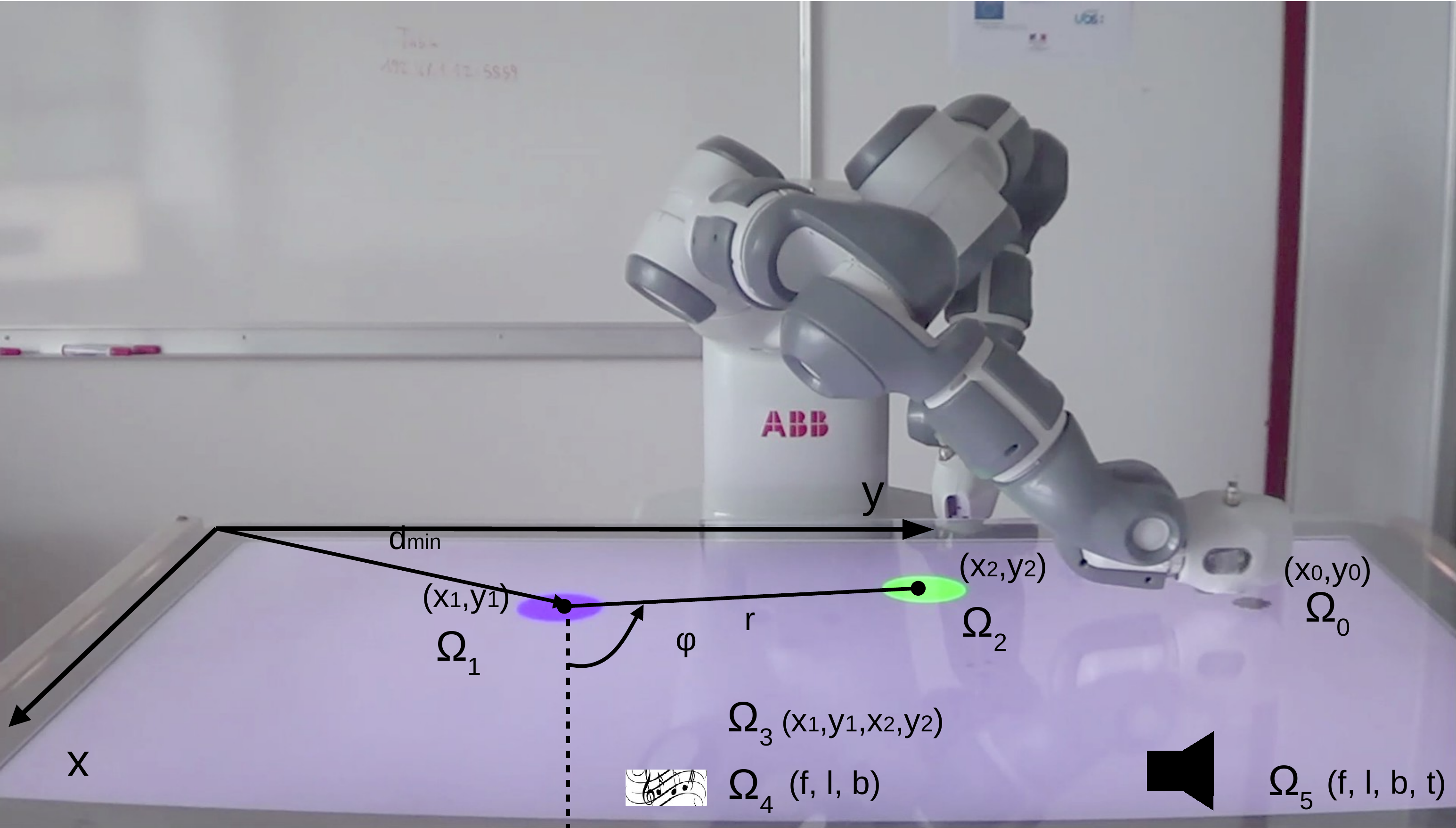}
\caption{Real Yumi setup: the 7-DOF industrial robot arm can produce sounds by moving the blue and green objects and touching the table. See \url{https://youtu.be/6gQn3JGhLvs} for an example task.}
\label{fig:physical_yumi_setup}
\end{figure}

Let us consider a reinforcement learning (RL)~\cite{RLSutton} robot placed in an environment surrounded by objects, without external rewards, but with human experts' help. How can the robot learn multiple tasks such as  manipulating an object, at the same time as 
combining objects together or other complex tasks that requires multiple steps? 

In the case of tasks  with various complexities and dimensionalities,  without a priori domain knowledge, the complexities of actions considered should be unbounded. 
 If we relate the complexity of actions to their dimensionality, actions of unbounded complexity should belong to spaces of unbounded dimensionality. For instance, if an action primitive of dimension $n$ is sufficient for placing an object to a position, 
a sequence of 2 primitives, i.e., an action of dimension $2n$ is sufficient to place the stick on 2 xylophone keys. 
 Nevertheless, tunes have variable lengths and durations. 
{Likewise, as a conceptual illustration, an unbounded sequence of actions is needed to control the interactive table and play tunes of any length in a setup like in Figure~\ref{fig:physical_yumi_setup}.} 
  In this work, we consider that actions of unbounded complexity can be expressed as  \textit{action primitives} and unbounded \textit{sequences of action primitives}, also named in~\cite{Zech2019IJRR} respectively \textit{micro} and \textit{compound actions}. 
The agent thus 
 needs to estimate the complexity of the task and deploy actions of the corresponding complexity.

To solve {this unbounded} problem, the learning agent should be \textit{starting small} 
before trying to learn more complex tasks as theorised in~\cite{Elman93}.  Indeed, in multi-task learning problems, some tasks can be compositions of simpler tasks (which we call \textit{`hierarchically organised tasks'}). This approach has been coined \textit{'curriculum learning'} in~\cite{Bengio2009P2AICML}.  The idea of this approach is to use the knowledge acquired from simple tasks to solve more complex tasks or high level of hierarchy tasks, or in other words, to leverage transfer learning (TL)~\cite{Pan2010ITKDE,Taylor2009JMLR,Weiss2016JD}. Uncovering the relationship between tasks is useful for transferring knowledge from a task to another. The insight behind TL is that generalization may occur not only within tasks, but also across tasks. 
 This is relevant 
  for compositional tasks. But how can the learning agent discover the decomposition of tasks and the relationship between tasks?
Moreover, transfer of knowledge between tasks can also be completed by transfer of knowledge from teachers. Indeed, humans and many animals do not just learn a task by trial and error. Rather, they extract knowledge about how to approach a problem from watching other people performing a similar task. Behavioural psychology studies~\cite{Whiten2000CS,Call2002Threesourcesof} highlight the importance of social and instructed learning, ``\textit{including learning about the consequences, sequential structure and hierarchical organisation of actions}''~\cite{Tomasello2007DS}.  Imitation is a mechanism for emerging representational capabilities~\cite{Piaget1952}. 
How can imitation enable the task decomposition into subtasks, and which kind of information should be transferred from the teacher to the learning agent to enable effective hierarchical reinforcement learning? How robust is this transfer to correspondence problems? How can teachers avoid demonstrations that could correspond to behaviours the agent already masters or require pre-requisites the robot has not learned yet? 
This work addresses \textbf{multi-task learning in open-ended environments by studying the role of transfer of knowledge across tasks} with the hypothesis that some tasks are interrelated, and the role of \textbf{ transfer of knowledge from other learners or experts} to determine how information is best transferred for  hierarchical reinforcement learning: when, what and whom to imitate? 

\section{State of the Art}

To learn unbounded sequences of motor actions for multiple tasks, we examine recent methods for curriculum learning based on intrinsic motivation. We  also review the methods for hierarchical reinforcement learning and imitation learning, which can be described as two types of transfer of knowledge.

\subsection{Intrinsic Motivation for Cross-Task Interpolation} 

In continual learning in an open-ended world without external rewards, to discover repertoires of skills, agents must be endowed with intrinsic motivation (IM), which is described in psychology as triggering curiosity in human beings~\cite{Deci85}, to explore the diversity of outcomes it can cause and to control its environment~\cite{Oudeyer2007ITEC,Schmidhuber2010ITAMD}. {These methods  use a reward function that is not shaped to fit a specific task but is general to all tasks the robot will face. Tending towards life-long learning, this approach, also called artificial curiosity, may be seen as a particular case of reinforcement learning using a reward function parametrised by internal features to the learning agent}. 
 One important form of IM system is the ability to autonomously set one’s own goals among the multiple tasks to learn. Approaches such as~\cite{Baranes2010,Rolf10a} have extended the heuristics of IM with goal-oriented exploration, and  proven to be able to learn fields of tasks in continuous task and action spaces of high but bounded dimensionality. More recently, IMGEP~\cite{Forestier2017C} and CURIOUS~\cite{Colas2019P3ICML} have combined intrinsic motivation and goal babbling with deep neural networks and replay mechanisms. They could select goals in a developmental manner from easy to more difficult tasks. Nevertheless, these works did not leverage cross-goal learning but only used interpolation between parametrised goals over a common memory dataset.

\subsection{Hierarchically Organised Tasks} 
Nevertheless, in the case of tasks  with various complexities and dimensionalities, especially with action spaces of unbounded dimensionality, those methods become intractable and the volume of the task and action spaces to explore grows exponentially. 
In that case, exploiting the relationships between the tasks can enable a learning agent to tackle increasingly complex tasks more easily, and heuristics such as social guidance can highlight these relationships.  
The idea would be to treat complex skills as assembly tasks, i.e., sequences of simpler tasks. This approach is in line with descriptions of motor behaviour of humans and primates as composing their early motions and being recombined after a maturation phase into sequences of action primitives~\cite{giszter2015motor}.  
In artificial systems, this idea has been implemented as a neuro-dynamic model by composing action primitives in~\cite{Arie2012RAS} and has been proposed in~\cite{Riedmiller2018P3ICML} to learn subtask policies and a scheduler to switch between subtasks, with offline off-policy learning, to derive a solution that is time-dependent on a scheduler. On the other hand, options were proposed  as a temporally abstract representation of complex actions made of lower-level actions and revealed faster to reach interesting subspaces as reviewed in~\cite{barto_behavioral_2013}. Learning simple skills then combining them by skill chaining is shown in~\cite{Konidaris2009ANIPSN} more effective than learning the sequences directly. Other approaches using temporal abstraction and hierarchical organization have been proposed~\cite{Barto2003DEDS}. 
  
More recently, Intrinsic Motivation has also tacked hierarchical RL to build increasingly complex skills by discovering and exploiting the task hierarchy using planning methods~\cite{Manoury2019HAI}. However it does not model explicitly a representation of the task hierarchy, letting planning compose the sequences in the exploitation phase. 
IM has also been used with temporal abstraction and deep leaning  in h-DQN~\cite{Kulkarni2016ANIPS} with meta-level learning subgoals and controller level policies over atomic actions. 
 However h-DQN was only applied to discrete state and action spaces. A similar idea has been proposed for continuous action and state spaces in {\cite{Duminy2018IIRC}, where the algorithm IM-PB} relies on a representation, called \textit{procedure}, of the task decomposition. A fully autonomous intrinsically motivated learner successfully discovers and exploits the task hierarchy of its complex environment, while still building sequences of action primitives of adapted sizes. These approaches could generalise across tasks by re-using the policies of subgoals for more complex tasks, once the task decomposition is learned. We would like to investigate here for continuous action and state spaces, the role 
transfer of  knowledge on task decomposition in hierarchical reinforcement learning, when there are several levels of hierarchy; and how transfer learning operates when tasks are hierarchically related compared to when tasks are similar.

\subsection{Active Imitation Learning (Social Guidance)}

Imitation learning techniques or Learning from Demonstrations (LfD)~\cite{Schaal1997ANIPS,Billard2007RobotProgrammingby} provide human knowledge for complex control tasks.  
However, imitation learning is often limited by the set of demonstrations. To overcome sub-optimality and noisiness of demonstrations, imitation learning has recently been combined with RL exploration, for instance when  initial human demonstrations have successfully initiated RL in~\cite{muelling2010learning,reinhart2017autonomous}. In~\cite{Taylor20111ICAAMSV} combination of transfer learning, learning from demonstration and reinforcement learning significantly improve both learning time and policy performance for a single task.
However, in the previously cited works, two hypotheses are frequently made: 
\begin{itemize}


\item the transfer of knowledge needs only one type of information. However, for multi-task learning, the demonstrations set should provide different types of information depending on the task and the knowledge of the learning. Indeed, in imitation learning works, different kinds of information for transfer of knowledge have been examined separately, depending on the setup at hand: external reinforcement signals~\cite{Thomaz2008CS}, demonstrations of actions~\cite{Grollman2010IRS}, demonstrations of procedures~\cite{Duminy2019FN}, advice operators~\cite{Argall2008} or disambiguation among actions~\cite{Chernova2009JAIR}.  The combination of different types of demonstrations has been studied in~\cite{Nguyen2012PJBR} for multi-task learning, where the proposed algorithm Socially Guided Intrinsic Motivation with Active Choice of Teacher and Strategy (SGIM-ACTS) showed that imitating a demonstrated action and outcome has different effects depending on the task, and that the combination of different types of demonstrations with autonomous exploration bootstraps the learning of multiple tasks. For hierarchical RL,  algorithm Socially Guided Intrinsic Motivation with Procedure Babbling  (SGIM-PB) in~\cite{Duminy2019FN} could also take advantage of demonstrations of actions and task decomposition. We propose in this study to examine the role of each kind of demonstrations with respect to the control tasks in order to learn task~hierarchy.
\item  the timing of these demonstrations has no~influence. 
 However, in curriculum learning the timing of knowledge transfer should be essential. Furthermore, the agent best knows when and what information it needs from the teachers, and active requests for knowledge transfer should be more efficient. 
 For instance, a reinforcement learner choosing when to request social guidance was  shown in~\cite{Cakmak2010AMDIT} making more progress. 
Such techniques are called active imitation learning or interactive learning, and echo the psychological descriptions of infants' selectivity in social partners and its link to their motivation to learn~\cite{Begus2018ALFICSMCLMCuriousLearners:How,Poulin-Dubois2011IBD}. 
Active imitation learning has been implemented~\cite{Fournier2019ITCDS} where the agent learns when to imitate using intrinsic motivation for a hierarchical RL problem in a discrete setting. For continuous action, state and goal spaces, the SGIM-ACTS algorithm~\cite{Nguyen2012PJBR} 
 uses intrinsic motivation to choose not only the kind of demonstrations, but also when to request for demonstrations and who to ask among several teachers. 
 SGIM-ACTS was extended for hierarchical reinforcement learning with the algorithm Socially Guided Intrinsic Motivation with Procedure Babbling  (SGIM-PB) in~\cite{Duminy2019FN}. In this article, we will study  whether a transfer of a batch of data or an active learner is more efficient to learn task hierarchy.  

\end{itemize}

\subsection{Summary: Our Contribution}


We combine both types of transfer of knowledge, across tasks and from experts, to address multi-task learning in a hierarchical RL problem in a non-rewarding, continuous and unbounded environment, {where experts with their own field of expertise, unknown to the robot, are available at the learner's request.} 
We propose to continue the approach initiated in~\cite{Duminy2019FN,Duminy2018ICSC} with the SGIM-PB algorithm that \textbf{combines intrinsic motivation, imitation learning and transfer learning} 
 to enable a robot to learn its curriculum by:

\begin{itemize}
	\item Discovering and exploiting the task hierarchy using a dual representation of complex actions in action and outcome spaces;
	\item Combining autonomous exploration of the task decompositions with imitation of the available teachers, using demonstrations as task dependencies;
	\item Using intrinsic motivation, and more precisely its empirical measures of progress, as its guidance mechanism to decide which information to transfer across tasks; and for imitation, when how and from which source of information to transfer.
\end{itemize}

In this article, we examine how task decomposition can be learned and transferred from a teacher or another learner using the mechanisms of intrinsic motivation in autonomous exploration and active imitation learning  for discovering task hierarchy for cross-task and cross-learner transfer learning.
More precisely, while in~\cite{Duminy2019FN} we showed on a toy simulation faster learning and better precision in the control, in this article, 
  we show on an industrial robot that \textbf{task decomposition is pivotal to completing tasks of higher complexity} (by adding more levels of hierarchy in the experimental setup), and we test the properties of our \textbf{active imitation of task decomposition : it is valid for cross-learner transfer even in the case of different embodiments, and active imitation proves more efficient than imitation of a batch dataset given from initialisation}.
  {This use-case study enables} deeper analysis into the mechanisms of the transfer of knowledge. 


%

The article is organized as follows: we describe our approach in Section~\ref{sec:approach}; and present our setups on the physical simulator and the real robot of an industrial robot arm in \mbox{Section~\ref{sec:exp}}. The results are analysed in Section~\ref{sec:results} and discussed in Section~\ref{sec:discussion}; finally we conclude this article in Section~\ref{sec:conclusion}.

\section{Our Approach\label{sec:approach}}


Grounding our work in cognitive developmental robotics~\cite{Asada2009ITAMD,Cangelosi2015}, we propose an intrinsically motivated learner able to self-organize its learning process for multi-task learning of hierarchically organized tasks by exploring action, task and task decomposition spaces. Our proposed algorithm combines autonomous exploration with active imitation learning into a learner discovering the task hierarchy to reuse its previously gained skills for tackling more complex ones, while adapting the complexity of its actions to the complexity of the task at hand.

In this section, we first formalize the learning problem we are facing. Then we describe the algorithm Socially Guided Intrinsic Motivation with Procedure Babbling (SGIM-PB). This algorithm uses a task-oriented representation of task decomposition called procedures to build more and more complex actions, adapting to the difficulty of the task.

\subsection{Problem Formalization}

Let us consider a robot, able to perform motions through the use of \textit{action primitives} $\pi^{\theta} \in \Pi $.
We suppose that the action primitives are parametrised functions with parameters of dimension $n$.  We note the parameters $\theta \in \mathbb{R}^{n}$. The action primitives represent the smallest unit of motions available to the robot.
The robot can also chain multiple action primitives together to form \textit{sequences of action primitives of any size $k \in \mathbb{N}$}.
We consider that the robot can execute actions in the form of a sequence of one or several action primitives. We note $\pi$  an action and will precise the parameter $\theta$ in the case of an action primitive $\pi^{\theta}$. We note $\Pi^\mathbb{N}$ the complete space of actions of any size available to the learner.

The environment can change as a consequence of the motions of the robot.
We call outcomes $\omega \in \Omega$  
these consequences. 
 They can be of various types and dimensionalities, and are therefore split in outcome subspaces $\Omega_i \subset \Omega$. Those outcomes can also be of different complexities, meaning that the actions generating these outcomes may require different numbers of action primitives to chain. The robot aims for  learning generalisation (how to reach a range of outcomes as broad as possible), and learning speed. It 
 learns which action to perform depending on the outcomes to generate, known as the inverse model $M : \omega \mapsto \pi$. A \textit{task} is thus a desired outcome, and the inverse model indicates which action can reach it. {As more than one action can lead to the same outcome, M is not a function. }


We take the trial and error approach, and we
 suppose that the error can be evaluated and $\Omega$ is a metric space, which means the learner can evaluate a distance between two outcomes $d(\omega_1, \omega_2)$.

\subsection{Procedures}

\begin{figure}[H]
\includegraphics[width=0.9\linewidth]{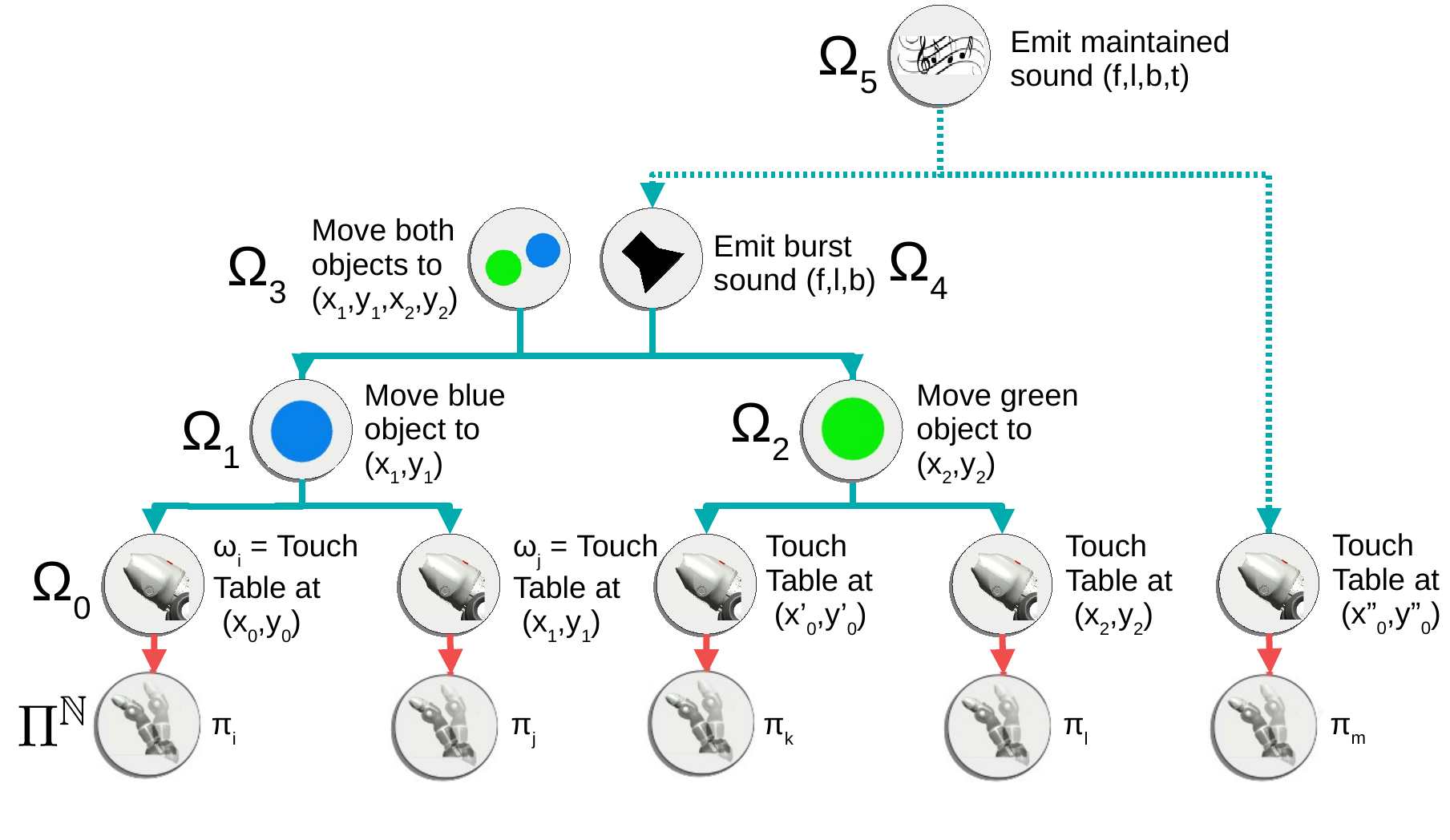}
\caption{Task hierarchy of the Yumi experimental setup : blue lines represent task decomposition into procedures for the simulation setup, dashed lines for the physical setup,  red lines for the direct inverse model. Eg. to move a blue object placed initially at $(x_0,y_0)$ to a desired position $(x_1,y_1)$, the robot can carry out task ($\omega_i$) that consists in moving its end-effector to $(x_0, y_0)$ to pick it, then task ($\omega_j$) that consists in moving the end-effector to $(x_1, y_1)$. These subtasks are executed  with respectively action primitives $\pi^i$  and $\pi^j$. Therefore to move the object, the learning agent can use the sequence of action primitives $(\pi^i,\pi^j)$. }
\label{fig:hierarchy}
\end{figure}

{Let us note $\mathcal{H}$ the hierarchy of the tasks used by our robot. $\mathcal{H}$ is formally defined as a directed graph where each node is a task $T$.  Figure~\ref{fig:hierarchy} shows a representation of task hierarchy}. 
As our algorithm is tackling the learning of complex hierarchically organized tasks, exploring and exploiting this hierarchy could ease the learning of the most complex~tasks.

To represent and learn this task hierarchy, we are using the procedure framework. This representation has been created as a way to push the learner to combine previously learned tasks to form more complex ones. A procedure is a combination of outcomes $(\omega_1, \omega_2) \in \Omega^{{2}}$. Carrying out a procedure $(\omega_1, \omega_2)$ means  
executing sequentially each component $\pi_{1,2}$ of the action sequence, where $\pi_i$ is an action that reaches $\omega_i$ $\forall i \in \llbracket 1, 2  \rrbracket$.

We can formalise task decomposition as a mapping from a desired outcome $\omega_g$ to a procedure $P: \omega_g \mapsto (\omega_1, \omega_2)$. $\omega_1$ and  $\omega_2$ are then called \textit{subtasks}. {In the task hierarchy $\mathcal{H}$, an outcome represents a task node in the graph, while the task decomposition represents the directed edges and the procedure is the list of its successors. $\mathcal{H}$ is initialised as a densely connected graph, and the exploration of SGIM-PB prunes the connections by testing which procedures or task decompositions respect the ground truth.} The procedure space or task decomposition space $\Omega^{{2}}$ is a new space to be explored by the learner to discover and exploit task decompositions. Our proposition is to derive the inverse model $M$ by using recursively the inverse model $M$ and the task decomposition $P$ until we derive a sequence of action primitives following this recursion:


\begin{equation*}
M(\omega_g) \mapsto \left\{
\begin{array}{rl}
 (\pi^{\theta_1},... \pi^{\theta_k})& \text{with } k \in \mathbb{N}\\
 (M(\omega_1), M(\omega_2) ) & \text{if } P(\omega_g) \mapsto (\omega_1, \omega_2)
 \end{array} \right.
\end{equation*}


\subsection{Algorithm} 

We here present the algorithm SGIM-PB (Socially Guided Intrinsic Motivation with Procedure Babbling).  In Section~\ref{sec:tlSetup}, we will also look at the variant named SGIM-TL (Socially Guided Intrinsic Motivation with Transferred Lump), which is provided at initialisation with a dataset of transferred procedures and their corresponding reached outcomes:  $\{ (\omega_i, \omega_j), \omega_r\}$.  {The main differences between SGIM-PB and SGIM-TL are outlined and they are contrasted with former versions IM-PB and SGIM-ACTS in Table \ref{tab:DiffAlgo}.} SGIM-PB and SGIM-TL propose to learn both $M$ and $P$ simultaneously.

\clearpage
\end{paracol}
\nointerlineskip
    \begin{specialtable}[H]
    \tablesize{\scriptsize}
    \widetable
\caption{Differences between SGIM-PB, SGIM-TL, SGIM-ACTS and IM-PB.}
\label{tab:DiffAlgo}

\setlength{\cellWidtha}{\columnwidth/6-2\tabcolsep+0.0in}
\setlength{\cellWidthb}{\columnwidth/6-2\tabcolsep+0.0in}
\setlength{\cellWidthc}{\columnwidth/6-2\tabcolsep-0.0in}
\setlength{\cellWidthd}{\columnwidth/6-2\tabcolsep-0.0in}
\setlength{\cellWidthe}{\columnwidth/6-2\tabcolsep-0.0in}
\setlength{\cellWidthf}{\columnwidth/6-2\tabcolsep-0.0in}
\scalebox{1}[1]{\begin{tabularx}{\columnwidth}{>{\PreserveBackslash\raggedright}m{\cellWidtha}>{\PreserveBackslash\raggedright}m{\cellWidthb}>{\PreserveBackslash\raggedright}m{\cellWidthc}>{\PreserveBackslash\raggedright}m{\cellWidthd}>{\PreserveBackslash\raggedright}m{\cellWidthe}>{\PreserveBackslash\raggedright}m{\cellWidthf}}
\toprule
\textbf{Algorithm} & \textbf{Action Representation} & \textbf{Strategies} \boldmath{$\sigma$} & \textbf{Transferred Dataset} & \textbf{Timing of the Transfer} & \textbf{Transfer of Knowledge} \\
\midrule
IM-PB~\cite{Duminy2018IIRC} &  parametrised actions, procedures & auton. action space explo., auton. procedural space explo. & None & NA & cross-task transfer \\
\midrule
SGIM-ACTS~\cite{Nguyen2012PJBR} &  parametrised actions & auton. action space explo., mimicry of an action teacher  & Teacher demo. of actions & Active request by the learner to the teacher & 
imitation \\
\midrule
SGIM-TL &  parametrised actions, procedures & auton. action space explo., auton. procedural space explo., mimicry of an action teacher, mimicry of a procedure teacher  & Another robot's procedures, Teacher demo. of actions and procedures & Procedures transf. at initalization time, Active request by the learner to the teacher & cross-task transfer, imitation
\\
\midrule
SGIM-PB &  parametrised actions, procedures & auton. action space explo., auton. procedural space explo., mimicry of an action teacher, mimicry of a procedure teacher  & Teacher demo. of actions and procedures & Active request by the learner to the teacher & cross-task transfer, imitation \\
\bottomrule
\end{tabularx}}
    \end{specialtable}
\begin{paracol}{2}
\switchcolumn

SGIM-PB starts learning from scratch. It is only provided with :

\begin{itemize}
	\item Dimensionality and boundaries of the action primitive space $\Pi$;
	\item Dimensionality and boundaries of each of the outcome subspaces $\Omega_i \subset \Omega$;
	\item Dimensionality and boundaries of the procedural spaces $(\Omega_i, \Omega_j) \subset \Omega^2$, defined as all possible pairs of two outcome subspaces.
\end{itemize}

Our SGIM-PB agent is to {collect data in order to learn how to reach all outcomes by generalising} from the sampled data. This means it has to learn for all reachable outcomes, the actions or procedures to use to reach the outcome. This corresponds to learning the inverse model $M$.  The model uses a local regression based on the k-nearest neighbours from the data collected by exploration. 
 In order to do that, the learning agent is provided with different {exploration \textit{strategies} (see Section~\ref{sec:ExplorationStrategies})}, {that are defined as methods to generate a procedure or action for} any given outcome. 
{The 4 types of strategies available to the robot are: two autonomous exploration strategies  and two interactive strategies per task type. For the autonomous exploration strategies, we consider action space exploration and procedural space exploration. For the interactive strategies, we consider mimicry of an action or a procedure teacher of the task type: the former's demonstrations are motor actions, while the latter's are procedures.}

 As these strategies could be more appropriate for some {tasks than others}, and as their effectiveness can depend on the maturity of the learning process, our learning agent needs to map the outcome subspaces and regions to the best suited strategies to learn them. {Thus, SGIM-PB uses an \textit{Interest Mapping} (see Section~\ref{sec:interest}) that associates to each strategy and region of an outcome space partition an \textit{interest} measure, so as to guide the exploration by indicating which tasks are the most interesting to explore at the current learning stage, and which strategy is the most efficient}.

The SGIM-PB algorithm (see Algorithm~\ref{sgim_hl}, Figure~\ref{fig:sgim_hl}) learns by episodes, and starts each episode by selecting an outcome  $\omega_g \in \Omega$ to focus on and an exploration strategy $\sigma$. 
The strategy and outcome region are selected {based on the Interest Mapping} by roulette wheel selection also called fitness proportionate selection, where the interest measure serves as fitness {(see Section~\ref{sec:interest})}. 

\begin{figure}[H]
\includegraphics[width=0.95\linewidth]{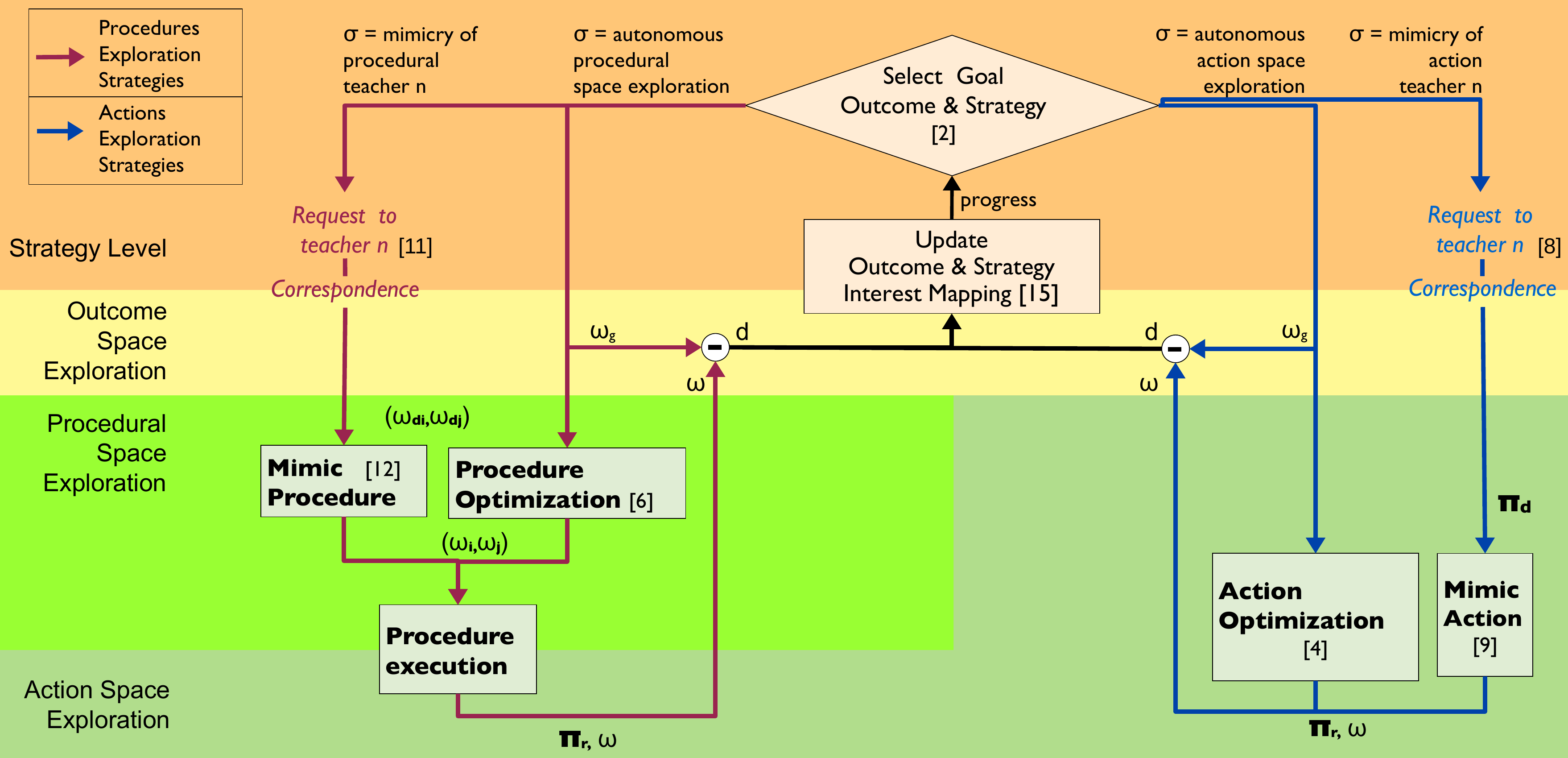}
\caption{SGIM-PB architecture: the arrows show the data transfer between the different blocks, the numbers between brackets refer to specific line number in Algorithm~\ref{sgim_hl}.}
\label{fig:sgim_hl}
\end{figure}
\vspace{-6pt}

\begin{algorithm}[H]
\setlength\baselineskip{15pt}
    \caption{SGIM-PB and its SGIM-TL variant}
    \label{sgim_hl}
    \begin{algorithmic}[1]
        \REQUIRE the different strategies $\sigma_1,...,\sigma_m$
        \ENSURE partition of outcome spaces $R \gets \bigsqcup_i \lbrace\Omega_i\rbrace$
        \REQUIRE \resubmit{CASE SGIM-TL: transfer dataset $D_3 =\{ (\omega_i, \omega_j), \omega_r\}$}
        \ENSURE \resubmit{CASE SGIM-TL: episodic memory \textit{Memo} $\gets \{ (\omega_i, \omega_j), \omega_r\}$}
        \LOOP
        \STATE $\omega_g, \sigma \gets$ Select Goal Outcome and Strategy($R$) \label{sgim_hl:choice}
        \IF{$\sigma$ = Autonomous action space exploration strategy} \label{sgim_hl:apol}
            \STATE \textit{Memo} $\gets$ Goal-Directed Action Optimization($\omega_g$)
        \ELSIF{$\sigma$ = Autonomous procedural space exploration strategy} \label{sgim_hl:aproc}
            \STATE \textit{Memo} $\gets$ Goal-Directed Procedure Optimization($\omega_g$)\label{sgim_hl:procoptim}        
        \ELSIF {$\sigma$ = Mimicry of action teacher $i$ strategy} \label{sgim_hl:mpol}
            \STATE $(\pi_d, \omega_d) \gets$ ask and observe demonstrated action to teacher $i$
            \STATE \textit{Memo} $\gets$ Mimic Action($\pi_d$)          
         \ELSIF{$\sigma$ = Mimicry of procedural teacher $i$ strategy} \label{sgim_hl:mproc}
            \STATE $((\omega_{di}, \omega_{dj}), \omega_d) \gets$ ask and observe demonstrated procedure to teacher $i$
            \STATE \textit{Memo} $\gets$ Mimic Procedure($(\omega_{di}, \omega_{dj})$)
        \ENDIF
        \STATE Update $M$ with collected data \textit{Memo}
        \STATE $R \gets$ Update Outcome and Strategy Interest Mapping($R$,\textit{Memo},$\omega_g$) \label{sgim_hl:interest}
        \ENDLOOP
    \end{algorithmic}
\end{algorithm}

At each episode, the robot and the environment are reset to their initial states. The learner uses the chosen strategy to build an action (based or not on a procedure), {decomposed into action primitives which} are then executed sequentially without getting back to its initial position. Whole actions are recorded, along with their outcomes. Each step (after each action primitive execution) of the sequence of action primitives execution is also recorded in $Memo$. 

\subsubsection{Exploration Strategies}
\label{sec:ExplorationStrategies}
In an episode under the autonomous action space exploration strategy, the learner tries to optimize the action $\pi$ to produce $\omega_g$ by stochastically choosing between random exploration of actions and  local regression on the k-nearest {action sequence neighbours. The probability of choosing local regression over random actions is proportional to the distance of $\omega_g$ to its nearest  action neighbours}. This action optimization is called Goal-Directed Action Optimization and based on the SAGG-RIAC algorithm~\cite{Baranes2010}.  

In an episode under the autonomous procedural space exploration strategy, the learner tries to optimize the procedure $(\omega_i, \omega_j) \in \Omega^2$ to produce $\omega_g$, by stochastically choosing between random exploration of the procedural space and local regression on the k-nearest {procedure neighbours.  The probability of choosing local regression over random actions is proportional to the distance of $\omega_g$ to its nearest procedure neighbours}. This process is called Goal-Directed Procedure Optimization. 

In an episode under the mimicry of an action teacher strategy, the learner requests a demonstration of an action to reach  $\omega_g$ from the chosen teacher. The teacher selects the demonstration $\pi_d$  as the action in its demonstration repertoire reaching the closest outcome from $\omega_g$.  It has direct access to the parameters of $\pi_d =(\pi_{d1},\pi_{d2},\ldots, \pi_{dl})$, and  explores locally the action parameters space (we do not consider the correspondence problem from teachers' demonstrations
). 

In an episode under the mimicry of a procedural teacher strategy, the learner requests a task decomposition of  $\omega_g$ from the chosen teacher. The demonstrated procedure $(\omega_{di}, \omega_{dj})$ will define a locality in the procedure space for it to explore.


When performing nearest neighbour searches during the execution of autonomous action and procedure exploration strategies (for local optimization of procedures or actions or when executing a procedure), the algorithm uses a performance metric which takes into account the complexity of the underlying action selected: 
\begin{equation}
perf(\omega_g) = d(\omega, \omega_g) \gamma^n
\end{equation}
where $d(\omega, \omega_g)$ is the normalized Euclidean distance between the target  and the reached outcomes $\omega_g$ and $\omega$, $n$ is the size of the action chosen (the length of the sequence of primitives), {$\gamma > 1$ is a constant used to balance accuracy and complexity of the action. In our case, we want the learner to mainly focus on accuracy while only avoiding overcomplex actions whenever possible, so we manually set a small $\gamma = 1.2$ very close to the lower limit. 

\subsubsection{Interest Mapping \label{sec:interest}}

After each episode, the learner stores the attempted procedures and actions, along with their reached outcomes in its memory. It then computes its competence in reaching the goal outcome $\omega_g$ by computing the distance $d(\omega_r, \omega_g)$ with the outcome $\omega_r$ it actually reached (if it has not reached any outcome in $\Omega_i$, we use a default value $d_{thres}$). 
Then interest measure is computed for the goal outcome and all outcomes reached during the episode (including the outcomes from a different subspace than the goal outcome):
\begin{equation}
interest(\omega, \sigma) = p(\omega)/K(\sigma)
\end{equation}
 where the progress $p(\omega)$ is the difference between the best competence for $\omega$ before and after the episode, $K(\sigma)$ is the cost associated to each strategy. $K(\sigma)$ is a meta parameter to favour some strategies such as autonomous ones, to push the robot to rely on itself as much as possible instead of bothering teachers. 

The interest measures are then used to partition the outcome space $\Omega$. The trajectory of the episode is added to their partition with hindsight experience replay (both goal and reached outcomes are taken into account), storing the values of the strategy $\sigma$, the outcome parameter, and the interest measure.  When the number of outcomes added to a region exceeds a fixed limit, the region is split into two regions with a clustering boundary that separates outcomes with low from those with high interest. This method is explained in more details in~\cite{Nguyen2014AR}.  The interest mapping is a tool to identify the zone of proximal development where the interest is maximal, and to organize the learning process. 


This interest mapping is initialized with the partition composed of each outcome type $\Omega_i \subset \Omega$. For the first episode, the learning agent always starts by choosing a goal and a strategy at random.

{In particular, this interest mapping enables SGIM-PB to uncover the task hierarchy by associating goal tasks and procedures. When testing a specific procedure $(\omega_i, \omega_j) \in \Omega^2$ that produces $\omega$ instead of the goal $\omega_g$ under the procedural space exploration or the mimicry of a procedure teacher strategies, SGIM-PB assesses the performance of this task decomposition and records the trajectory of the episode in the memory. This task decomposition is likely to be reused again during local regression of k-nearest neighbours for tasks close to $\omega$ and $\omega_g$ and for short sequences of primitives, i.e., if its performance $perf(\omega_g) = d(\omega, \omega_g) \gamma^n$ is high. Thus the different task decompositions are compared both for their precision and cost in terms of complexity. At the same time, SGIM-PB updates the interest map for that strategy. If the procedure is  not relevant for the goal $\omega_g$, the procedure is ignored henceforward. On the contrary if the procedure is the right task decomposition, the interest $interest(\omega, \sigma)$ for this procedure exploration/mimicry strategy $\sigma$ increases. Thus, conversely, SGIM-PB continues to explore using the same strategy, and tests more procedures for the same region of outcomes. As the number of procedures explored increases and are selected by intrinsic motivation (using the performance and interest criteria), SGIM-PB  associates goal tasks to the relevant procedures, hence builds up the adequate task decomposition. These associations to procedures constitute the task hierarchy uncovered by the robot.}

\section{Experiment \label{sec:exp}}

\rmk{We show that (1) imitation learning of procedures (SGIM-PB vs IM-PB) bootstrap the learning for 2nd level hierarchical tasks or higher ;  (2) procedures are essential to learn fro 4th level hierarchical tasks (SGIM-PB vs SGIM-ACTS); (3) these bootstrapping effects are robot to correspondence problems in the procedure spaces. Moreover active imitation learning is efficient compared to batch transfer of knowledge from another robot (SGIM-PB vs SGIM-TL).
Therefore, owing to the procedures representation, transfer of knowledge from teachers (experts or other robots) by imitation learning of actions and task decomposition benefits to hierarchical reinforcement learning because SGIM-PB performs better than SGIM-ACTS for high-level tasks. }

In this section, we present the experiment we conducted to assess the capability of our SGIM-PB algorithm. 
The experimental setup features the 7 DOF right arm of an industrial Yumi robot from ABB which can interact with an interactive table and its virtual objects. Figure~\ref{fig:physical_yumi_setup} shows the robot facing a
tangible RFID interactive table from~\cite{
kubicki2016using}. The robot learns to interact with it using its end-effector. It can learn an infinite number of hierarchically organized tasks regrouped in 5 types of tasks, using sequences of motor actions of unrestricted size.   The experiments have been carried out with the physical industrial robot ABB and with simulation software provided with the robot, Robotstudio. While the software provided by the robot can provide static inverse kinematics, it can not provide movement trajectories.

 We made preliminary tests of our SGIM-PB learner on a physical simulator of the robot. We will call this the \textit{simulation setup}. In the simulation setup, the end-effector is the tip of the vacuum pump below its hand. The simulation setup will be modified as a setup called \textit{left-arm setup} in Section~\ref{sec:tlSetup}.

We also implemented this setup for a physical experiment, to compare both interactive strategies more fairly using SGIM-ACTS and SGIM-PB. For that, we modified the procedural teacher strategy so they have the same limited repertoire from which to draw demonstrations as action teachers. We also added an extra more complex task without demonstrations to compare the autonomous exploration capability of both algorithms. The setup, shown on Figure~\ref{fig:physical_yumi_setup}, will be referred to as the \textit{physical setup}. In the physical setup, the end-effector is the bottom part of the hand.

In the following subsections, we describe in more details the physical setup and the simulation setup and their variables, while mentioning their differences. Then we end this section by presenting the teachers and evaluation methods.

\subsection{Experimental Setup \label{sec:setup}}

The position of the arm's tip on the table (see Figure~\ref{fig:table}) is noted $(x_0, y_0)$. Two virtual objects (disks of radius $R = 4$ cm) can be picked and placed, by placing the arm's tip on them and moving it at another position on the table. Once interacted with, the final positions of the two objects are given to the robot by the table, respectively $(x_1, y_1)$ and $(x_2, y_2)$.
Only one object can be moved at a time, otherwise the setup is blocked and the robot's motion cancelled. If both objects have been moved, a burst sound is emitted by the interactive table, parametrised by its frequency $f$, its intensity level $l$ and its rhythm $b$. It can be maintained for a duration $t$ by touching a new position on the table afterwards.
The sound parameters are computed with arbitrary rules so as to have both linear and non linear relationships, as follow:
\begin{align}
\vspace{-5cm}
f &= (D/4 - d_{min}) 4 / D \\
l &= 1 - 2 (ln(r) - ln(r_{min})) / (ln(D) - ln(r_{min})) \\
b &= (\vert \varphi \vert / \pi) * 0.95 + 0.05 \\
t &= d_2 / D
\end{align}
where $D$ is the interactive table diagonal, $(r, \varphi)$ are the polar coordinates of the green object in the system centred on the blue object, $r_{min} = 2 R$, $d_{min}$ is the distance between the blue object and closest table corner, and $d_2$ is the distance between the end effector position on the table and the green object . 

\begin{figure}[H]
\includegraphics[width=0.95\linewidth]{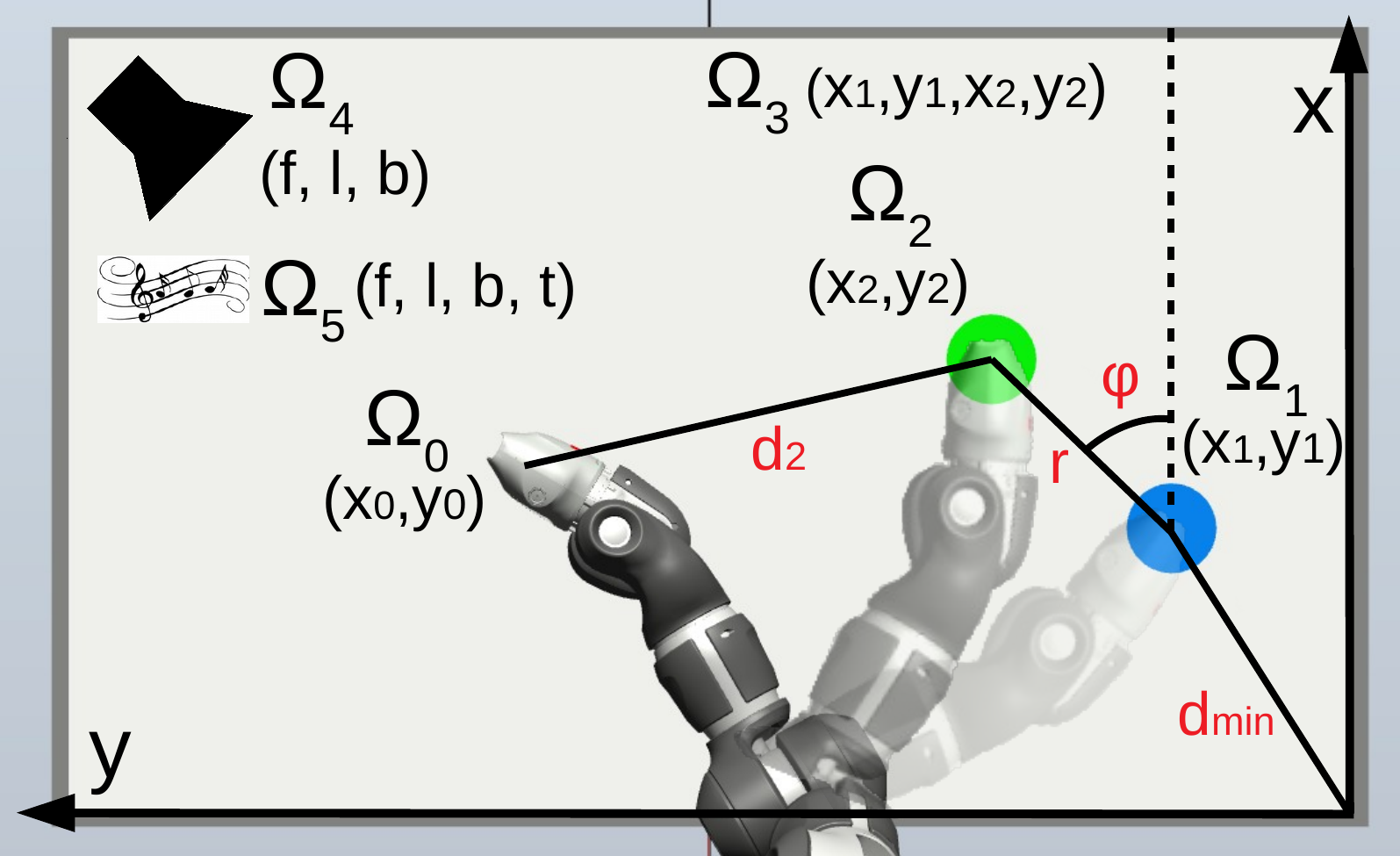}
\caption{Representation of the interactive table: the first object is in blue, the second one in green, the produced burst sound and maintained sound are also represented in top left corner. The outcome spaces are also represented on this figure with their parameters in black, while the sound parameters are in red.}
\label{fig:table}
\end{figure}

The interactive table state is refreshed after each {action primitive} executed. The robot is not allowed to collide with the interactive table. In this case, the motor action is cancelled and reaches no outcomes. Before each attempt, the robot is set to its initial position and the environment is reset. 

\subsection{Experiment Variables}

\subsubsection{Action Spaces \label{sec:yumi_policy}}

The motions of each joint are controlled by a one-dimensional Dynamic Movement Primitive (DMP)~\cite{pastor_learning_2009}. We will use the same notations for the DMP parameters, defined by the system:
\begin{align}
\tau \dot{v} &= K(g - x)-Dv+(g-x_{0})f(s) \\
\tau \dot{x} &= v \\
\tau \dot{s} &= -\alpha s
\end{align}
where $x$ and $v$ are the position and velocity of the system; $s$ is the phase of the motion; $x_0$ and $g$ are the starting and end position of the motion; $\tau$ is a factor used to temporally scale the system (set to fix the {duration} of an action primitive execution); $K$ and $D$ are the spring constant and damping term fixed for the whole experiment; $\alpha$ is also a constant fixed for the experiment; and $f$ is a non-linear term used to shape the trajectory called the forcing term. This forcing term is defined as:
\begin{equation}
f(s) = \frac{\sum\nolimits_{i} w_{i}\psi_{i}(s)s}{\sum\nolimits_{i} \psi_{i}(s)}
\end{equation}
where $\psi_{i}(s)=\exp(-h_{i}(s-c_{i})^2)$ with centers $c_i$ and widths $h_i$ fixed for all primitives. There are 1 weights $w_i$ per DMP, which is therefore simply noted $w$.

The weights of the forcing term and the end positions are the only parameters of the DMP used by the robot. One weight per DMP is used and each DMP controls one joint. The starting position of a primitive is set by either the initial position of the robot (if it is starting a new action) or the end position of the preceding primitive. Therefore an action primitive $\pi^{\theta}$ is parametrized by:
\begin{equation}\label{eq:yumi_policy}
\theta = (a_0, a_1, a_2, a_3, a_4, a_5, a_6)
\end{equation}
where $a_i = (w^{(i)}, g^{(i)})$ corresponds to DMP parameters of the joint $i$:  the final joint angle $g^{(i)}$, and the parameters $w^{(i)}$ of a basis function for the forcing term. The action primitive space is thus $\Pi = \mathbb{R}^{14}$, and the complete action space $(\mathbb{R}^{14})^{\mathbb{N}}$.


\subsubsection{Outcome Spaces}

The outcome spaces the robot learns are hierarchically organized:

\begin{itemize}
    \item $\Omega_0 = \{(x_0,y_0)\}$: positions touched on the table;
    \item $\Omega_1 = \{(x_1,y_1)\}$: positions of the first object;
    \item $\Omega_2 = \{(x_2,y_2)\}$: positions of the second object;
    \item $\Omega_3 = \{(x_1,y_1,x_2,y_2)\}$: positions of both objects;
    \item $\Omega_4 = \{(f,l,b)\}$: burst sounds produced;
    \item $\Omega_5 = \{(f,l,b,t)\}$: maintained sounds produced.
\end{itemize}

The outcome space is a composite and continuous space (for the physical setup $\Omega = \bigcup^5_{i=0} \Omega_i$, for the simulation setup $\Omega = \bigcup^4_{i=0} \Omega_i$), containing subspaces of 2 to 4 dimensions. Multiple interdependencies are present between tasks: controlling the position of either the blue object ($\Omega_1$) or the green object ($\Omega_2$) comes after being able to touch the table at a given position ($\Omega_0$); moving both objects ($\Omega_3$) or making a sound ($\Omega_4$) comes after being able to move the blue and the green object
, the maintained sound ($\Omega_5$) is the most complex task of the physical setup. This hierarchy is shown on Figure~\ref{fig:hierarchy}.

\textls[-20]{Our intuition is that a learning agent should start by making a good progress in the easiest task $\Omega_0$, then $\Omega_1$, $\Omega_2$. Once it mastered those easy tasks, it can reuse that knowledge to learn to achieve the most complex tasks $\Omega_3$ and $\Omega_4$. We will particularly focus on the learning of the $\Omega_4$ outcome space and the use of the procedure framework for it. Indeed in this setting, the relationship between a goal outcome in $\Omega_4$ and the necessary positions of both objects ($\Omega_1, \Omega_2$) to reach that goal are not linear. So with this setting, we test if the robot can \textbf{learn a non-linear mapping between a complex task and a procedural space}. Finally for the physical version, we see if the robot can reach and explore the most complex task $\Omega_5$ in the absence of an allocated teacher.}

\subsection{Teachers \label{sec:simu_yumi_teachers}}
\label{sec:teachers}
\textls[-20]{To help SGIM-PB in the simulation setup, procedural teachers (strategical cost $K(\sigma)=5$, compared to $K(\sigma)=1$ for autonomous strategies) were available for every outcome space except $\Omega_0$. Each teacher gives on the fly a procedure adapted to the learner's request, according to its domain of expertise and} 
 according to a construction rule:

\begin{itemize}
    \item ProceduralTeacher1: $\Omega_1 \to (\Omega_0^2)$; 
    \item ProceduralTeacher2: $\Omega_2 \to (\Omega_0^2)$; 
    \item ProceduralTeacher3: $\Omega_3 \to (\Omega_1, \Omega_2)$;
    \item ProceduralTeacher4: $\Omega_4 \to (\Omega_1, \Omega_2)$. 
\end{itemize}

We also added different 
action teachers (strategical cost of $K(\sigma)=10$), each expert of one outcome space:

\begin{itemize}
    \item ActionTeacher0 ($\Omega_0$): 11 demos of action primitives;
    \item ActionTeacher1 ($\Omega_1$): 10 demos of size 2 actions;
    \item ActionTeacher2 ($\Omega_2$): 8 demos of size 2 actions;
    \item ActionTeacher34 ($\Omega_3$ and $\Omega_4$): 73 demos of size 4 actions.
\end{itemize}

In the physical setup, we want to delve into the differences between action and procedural teachers. 
 So as to put them on an equal footing, we used for all teachers demonstration datasets of limited sizes. The demonstrations in the physical version of the action and procedural teachers reach the same outcomes for $\Omega_1, \Omega_2, \Omega_3$ and $\Omega_4$. 
An extra action teacher provides demonstrations for the simplest outcome space $\Omega_0$:

\begin{itemize}
    \item ActionTeacher0 ($\Omega_0$): 9 demos of action primitives;
    \item ActionTeacher1 ($\Omega_1$): 7 demos of size 2 actions;
    \item ActionTeacher2 ($\Omega_2$): 7 demos of size 2 actions;
    \item ActionTeacher3 ($\Omega_3$): 32 demos of size 4 actions;
    \item ActionTeacher4 ($\Omega_4$): 7 demos of size 4 actions;
\end{itemize}

The  demonstrations for the procedural teachers correspond to the subgoals reached by the action primitives of the action teachers. The procedural teachers have the same number of demonstrations as their respective action teachers. 

The action teachers were provided to the SGIM-ACTS learner, while the SGIM-PB algorithm had all procedural teachers and ActionTeacher0. No teacher was provided for the most complex outcome space $\Omega_5$, as to compare the autonomous exploration capability of both learners.

In both setups, the number of demonstrations was chosen arbitrarily small, and the higher the dimensionality of the outcome space it teaches, the more demonstrations it can offer. The demonstrations were chosen so as to cover uniformly the reachable outcome~spaces.

\subsection{Evaluation Method\label{sec:yumi_evaluation}}

\textls[-20]{To evaluate our algorithm, we created a testbench set of goals uniformly covering the outcome space (the evaluation outcomes are different from the demonstration outcomes). It has 29,200~goals for the real robot and 19,200 goals for simulated version. The evaluation consists in computing mean Euclidean distance between each of the testbench goals and their nearest neighbour in the learner memory ($d_{thres}$= 5).}
The evaluation is repeated regularly across the learning~process.

Then to assess the efficiency of our algorithm in the simulation setup, we are comparing the averaged results of 10 runs of 25,000 learning iterations \resubmit{(each run took about a week to proceed)} of the following algorithms :

\begin{itemize}
    \item RandomAction: random exploration of the action space $\Pi^{\mathbb{N}}$;
    \item IM-PB: autonomous exploration of the action $\Pi^{\mathbb{N}}$ and procedural space $\Omega^2$ driven by intrinsic motivation;
    \item SGIM-ACTS: interactive learner driven by intrinsic motivation. Choosing between autonomous exploration of the action space $\Pi^{\mathbb{N}}$ and mimicry of any action teacher;
    \item SGIM-PB: interactive learner driven by intrinsic motivation. Has autonomous exploration strategies (of the action $\Pi^{\mathbb{N}}$ or procedural $\Omega^2$ space) and mimicry ones for any procedural teacher and ActionTeacher0;
    \item Teachers: non-incremental learner only provided with the combined knowledge of all the action teachers.
\end{itemize}

For the physical setup, we are only comparing 1 run of SGIM-ACTS and SGIM-PB \resubmit{(20,000 iterations or one month of trials for each)}, as to delve deeper into the study of the interactive strategies and their impact. 

The codes used are available at \url{https://bitbucket.org/smartan117/sgim-yumi-simu} (simulated version), and at \url{https://bitbucket.org/smartan117/sgim-yumi-real} (physical~one).

\section{Results}
\label{sec:results}

\rmk{
(0) SGIM-PB learns to complete multiple tasks better than the other algorithms to learn task decomposition and to estimate the length of the action
(00) This is due to a self-devised curriculum of choice of tasks and sampling strategy
(1) SGIM-PB learns better than IM-PB : imitation learning bootstraps the learning of procedures
(2) SGIM-PB learns better than SGIM-ACTS : procedures are essential for higher level of hierarchy tasks.
(3) SGIM-PB learns from a left-hand procedural teacher : it is robust to correspondence problems in the procedural space (we showed that SGIM-ACTS is robust to correspondence problems in~\cite{Nguyen2014AR})
(4) SGIM-TL is not better than SGIM-PB : batch learning (a large amount of information from the beginning) is not more efficient than active learning
}


\subsection{Task Decomposition and Complexity of Actions}
\subsubsection{Task Hierarchy Discovered}

 To check whether SGIM-PB is capable of learning the task hierarchy used to build the setup (see Figure~\ref{fig:hierarchy}), after an exploration of 25,000 compound actions, we tested how it reaches each of the goals of the testbench. More precisely, to show the task hierarchy learned by the algorithm, we reported the types of procedures SGIM-PB uses to reach goals of each goal types and plot in Figure~\ref{fig:proc_task} the histogram of the procedures used for $\Omega_1$ and $\Omega_4$ for the simulated (left column) and for $\Omega_1, \Omega_4$ and $\Omega_5$ for physical setup (right column). 
  We can see that in both setups, the most associated procedural space with each outcome space corresponds to the designed task hierarchy, namely, $\Omega_1$ uses procedures mostly procedures $(\Omega_0,\Omega_0)$ and $\Omega_4$ uses $(\Omega_1,\Omega_2)$ and $\Omega_5$ uses $(\Omega_4,\Omega_0)$. It also corresponds to the hierarchy demonstrated by the procedural teachers. It was even capable to learn the task hierarchy in the absence of provided teacher for the  outcome space $\Omega_5$ in the physical setup. \textbf{The SGIM-PB learner was able to learn the task hierarchy of the setup using the procedure framework.} 
 The procedure repartition is not perfect though, as other subgoals are also used. For instance, procedures  $(\Omega_1, \Omega_2)$ were associated to the $\Omega_1$ goals. Although they correctly reach the goal tasks, they introduce longer sequences of actions and are thus suboptimal. 

\begin{figure}[H]
\centering
\minipage{\linewidth}
\minipage{0.38\linewidth}
\minipage{0.5\linewidth}
\includegraphics[width=\linewidth]{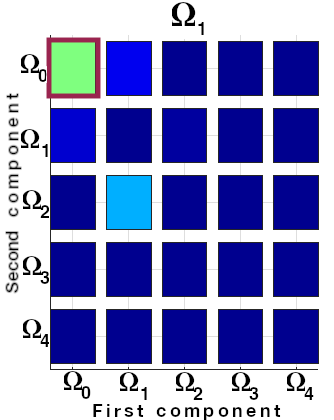}
\endminipage
\minipage{0.5\linewidth}
\includegraphics[width=\linewidth]{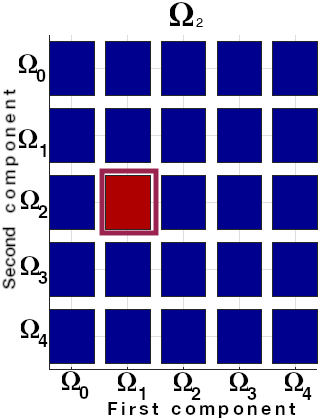}
\endminipage
\endminipage 
\hfill
\vline
\minipage{0.57\linewidth}
\minipage{0.33\linewidth}
\includegraphics[width=\linewidth]{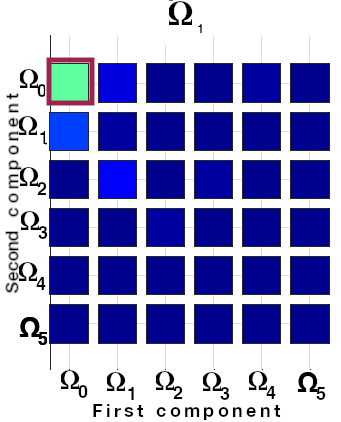}
\endminipage
\minipage{0.33\linewidth}
\includegraphics[width=\linewidth]{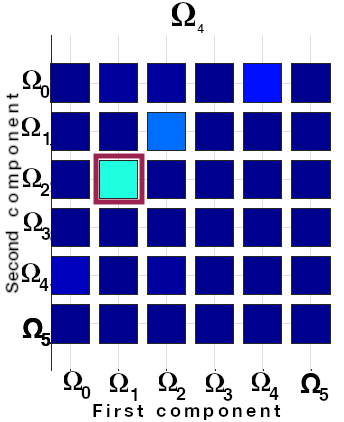}
\endminipage
\minipage{0.33\linewidth}
\includegraphics[width=\linewidth]{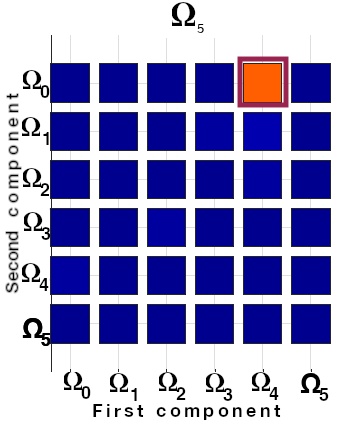}
\endminipage
\endminipage
\vline
\minipage{0.03\linewidth}
\includegraphics[width=\linewidth]{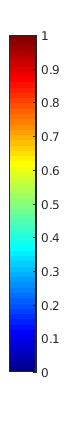}
\hfill
\endminipage

\minipage{\linewidth}
\caption{Task 
 hierarchy discovered by the SGIM-PB learners: this represents for 3 complex outcome space the percentage of time each procedural space would be chosen for the physical setup (left) 
 and the simulated setup (center). The percentage scale is represented in the colorbar (right). }
\label{fig:proc_task}
\endminipage
\endminipage
\end{figure}

  \subsubsection{Length of the Chosen Actions}

 Indeed, this internal understanding of task hierarchy translates into visible actions, in particular the length of action primitive sequences which should adapt to the level of hierarchy of the goal task. 
 We analysed which action size is chosen by the local action optimization function 
  of SGIM-PB, 
 for each goal task of the evaluation testbench. The results are shown on Figure~\ref{fig:nb_actions}
  for each outcome space. In both setups, \textbf{SGIM-PB was able to scale the complexity of its actions}: using mostly primitive and sequences of 2 primitives for $\Omega_0$;  sequences of 2~primitives  for $\Omega_1$ and $\Omega_2$;  of 4 primitives for $\Omega_4$ and $\Omega_3$; and  of 5 primitives for $\Omega_5$ on the physical setup. It is not perfect as SGIM-PB partially associated the $\Omega_0$ outcome space with 2-action primitives when single action primitives were sufficient. Because all tasks require the robot to touch the table, and thus has an outcome $\Omega_0$ on the environment, all the complex actions used for the other tasks could be associated to reach $\Omega_0$: the redundancy of this  model makes it harder to select the optimal action size.

\begin{figure}[H]
\includegraphics[width=0.95\linewidth]{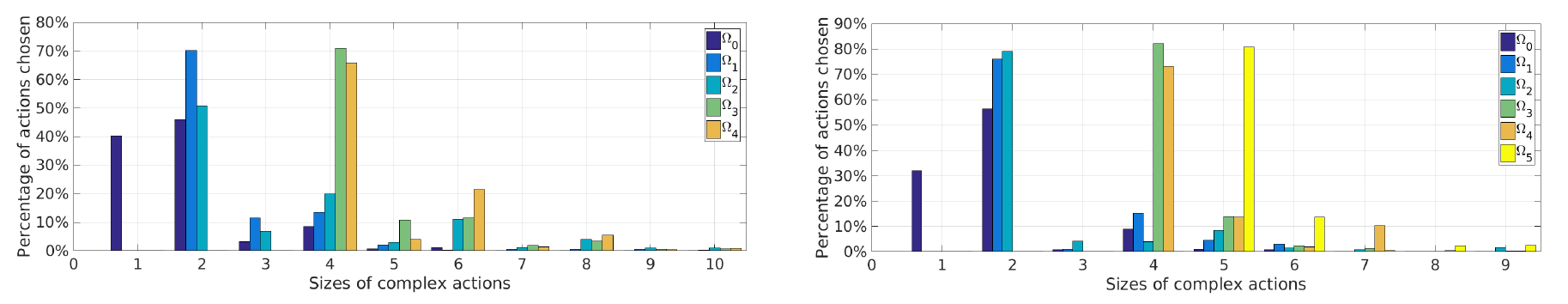}
\caption{Histogram of the action size of actions chosen by  SGIM-PB for each outcome space. Test results on the testbench tasks on the simulation (\textbf{left}) and physical setups (\textbf{right}).}
\label{fig:nb_actions}
\end{figure}

This understanding of task complexity and task hierarchy also leads to a better performance of SGIM-PB. Figure~\ref{fig:simu_yumi_evaluation} shows the global evaluation results of all tested algorithms for the simulated version. It plots the test mean error made by each algorithm to reach the goals of the benchmark with respect to the number of actions explored during learning. SGIM-PB has the lowest level of error compared to the other algorithms. 
 Thus \textbf{SGIM-PB learns with better precision}. This is due to its transfer of knowledge from simple tasks to complex tasks, after learning task hierarchy, owing to both its use of imitation learning and the procedure representation.

\begin{figure}[H]
\includegraphics[width=0.95\linewidth]{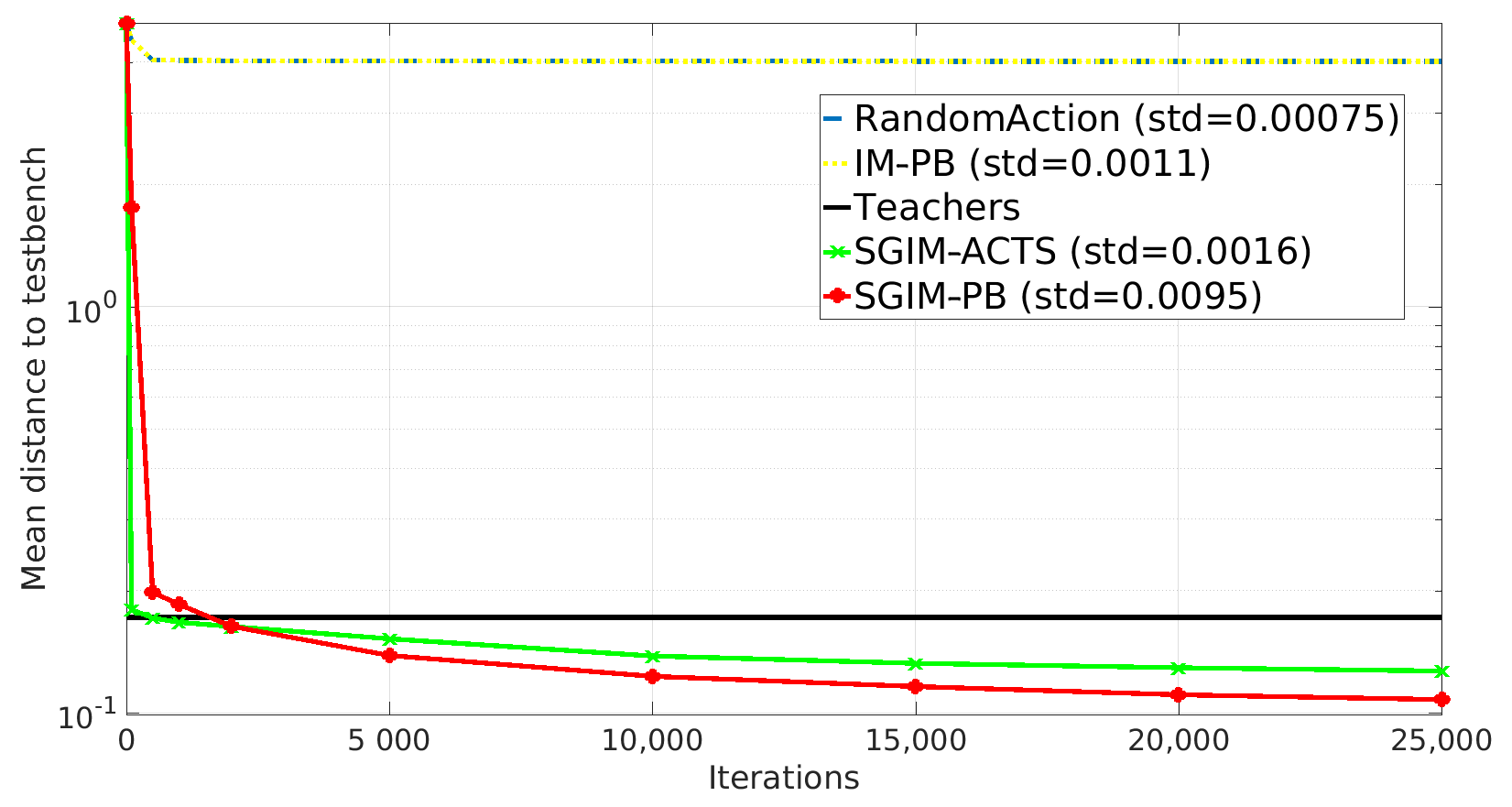}
\caption{Evaluation 
 of 
 all algorithms throughout the learning process for the simulation setup, final standard deviations are given in the legend.}
\label{fig:simu_yumi_evaluation}
\end{figure}

\subsection{Imitation Learning Bootstraps the Learning of Procedures: Role of Imitation}

First, let us examine the role of imitation by contrasting the algorithms with active imitation (SGIM-PB and SGIM-ACTS) with the autonomous learners without imitation (RandomAction and IM-PB). IM-PB is the variation of SGIM-PB with only autonomous learning, without imitation of actions or imitation of procedures. 
In  Figure~\ref{fig:simu_yumi_evaluation}, both autonomous learners 
have higher final levels of error than the active imitation learners
, which shows the advantage of using social guidance. Besides, both SGIM-ACTS and SGIM-PB have error levels dropping below that of the teachers, showing they learned further than the provided action demonstrations: the combination of autonomous exploration and imitation learning improves the learner's performance beyond the performance of the~teachers.

Figure~\ref{fig:simu_yumi_task_eval} plots the evaluation on the simulated setup for each type of tasks. While all algorithms have about the same performance on the simple task $\Omega_0$, we notice a significant difference for the complex tasks  in $\Omega_1, \Omega_2, \Omega_3$ or $\Omega_4$ between the autonomous 
  and the active imitation learners
  . The autonomous learners were not able to even reach a goal in the complex subspaces. In particular, the difference between IM-PB and SGIM-PB means that imitation 
   is necessary to learn complex tasks, it is not only a speeding-up effect. 

\begin{figure}[H]
\includegraphics[width=0.95\linewidth]{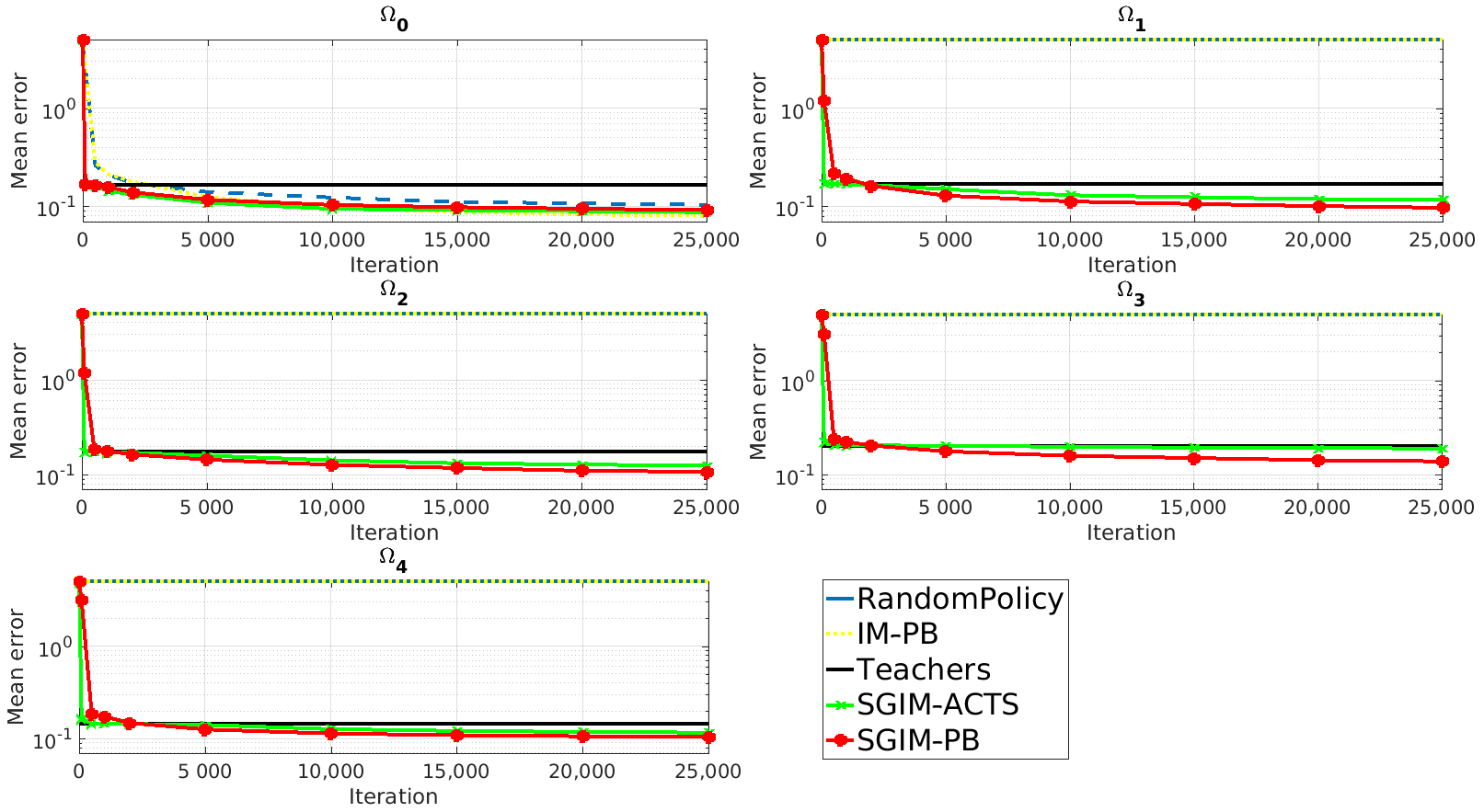}
\caption{Evaluation 
 of 
 all algorithms per outcome space on the simulation setup  (RandomAction and IM-PB are superposed on all except for $\Omega_0$).}
\label{fig:simu_yumi_task_eval}
\end{figure}
 
These results show that 
the imitation strategies have improved the performance from the beginning of the learning process, and this improvement is more visible for complex tasks. \textbf{More than a speeding effect, imitation enables the robot to reach the first goals in the complex task subspaces, to be optimised and generalised later on by autonomous exploration, so that SGIM-PB is not bound by the limitations of the demonstration~dataset}. 


\subsection{Procedures Are Essential for Higher Level of Hierarchy Tasks: Role of Procedures}
Second, let us examine the role of procedures, both for imitation and for autonomous exploration. 

\subsubsection{Demonstrations of Procedures} 
To analyse the difference between the different imitation strategies, i.e between imitation of action primitives and imitation of procedures,
we can compare the algorithms SGIM-PB and SGIM-ACTS. While SGIM-PB has procedures teachers for the complex tasks and can explore the procedure space to learn task decomposition, SGIM-ACTS has action teachers and does not have the procedure representation to learn task decomposition. Instead SGIM-ACTS explores the action space by choosing a length for its action primitive sequence, then the parameters of the primitives.

\textls[-20]{In Figure~\ref{fig:simu_yumi_evaluation}, we see that SGIM-PB is able to outperform SGIM-ACTS after only 2000~iterations, which suggests that procedural teachers can effectively replace action teachers for complex tasks. 
More precisely, as shown in Figure~\ref{fig:simu_yumi_task_eval}}, 
 for tasks $\Omega_0$ where SGIM-ACTS and SGIM-PB have the same action primitive teacher ActionTeacher0, there is no difference in performance, and SGIM-PB was outperforming SGIM-ACTS on all the complex tasks, particularly $\Omega_3$.

\textls[-20]{To understand the learning process of SGIM-PB that leads to this difference in performance, let us look at the evolution of the choices of each outcome space (Figure~\ref{fig:simu_yumi_tasks}) and strategy (Figure~\ref{fig:simu_yumi_strats}). The improvement of performance of SGIM-PB compared to the other algorithms can be explained in Figure~\ref{fig:simu_yumi_tasks} by its  choice of task $\Omega_3$ in the curriculum for iterations above 10,000 after an initial phase where it explores all outcome spaces.
\mbox{Figure~\ref{fig:simu_yumi_strats}}, we notice that SGIM-PB chooses mainly as strategies: ProceduralTeacher3 among all imitation strategies, and Autonomous procedures among the autonomous exploration strategies. The histogram of each task-strategy combination chosen for the whole learning process in \mbox{Figure~\ref{fig:simu_yumi_choices}} confirms that ProceduralTeacher3 was chosen the most frequently among imitation strategies specifically for tasks $\Omega_3$, but SGIM-PB used most extensively Autonomous procedures. Thus \textbf{SGIM-PB performance improvement is correlated with its procedure space exploration with both Autonomous procedures and procedural imitation strategies}.}

\begin{figure}[H]
\includegraphics[width=\linewidth]{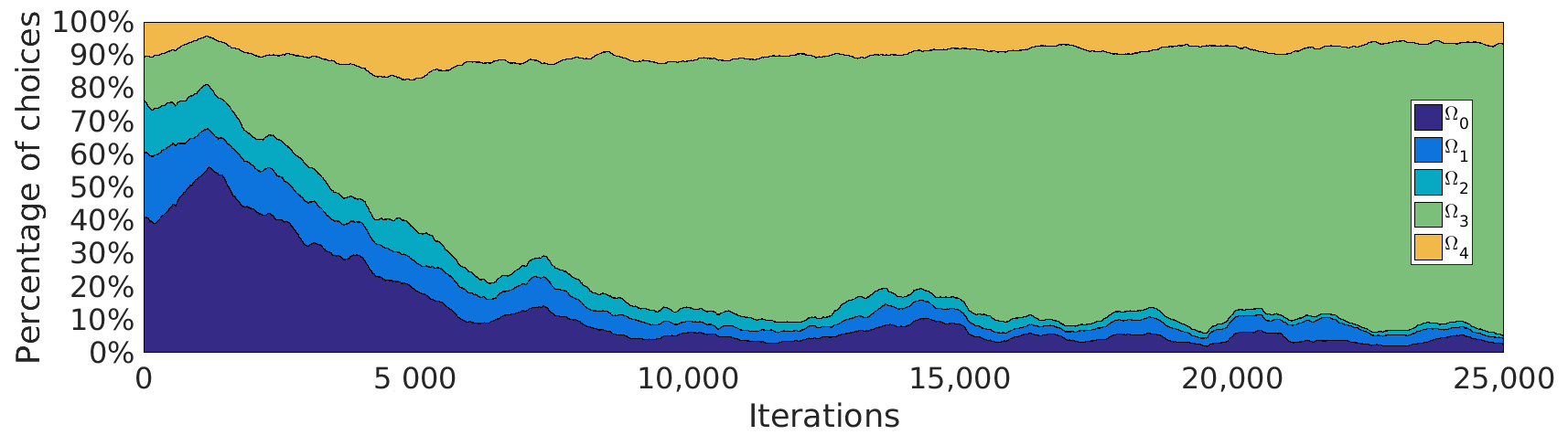}
\caption{{Evolution} 
 of choices of tasks for the SGIM-PB learner during the  learning process on the simulation setup.}
\label{fig:simu_yumi_tasks}
\end{figure}

This comparison between SGIM-PB and SGIM-ACTS  on simulation is confirmed on the physical setup. The global evaluation in Figure~\ref{fig:real_yumi_evaluation} shows a significant gap between the two performances. SGIM-PB  even outperforms SGIM-ACTS after only 1000 iterations, which suggests that procedural teachers can be more effective than action teachers for complex tasks. 

\begin{figure}[H]
\includegraphics[width=\linewidth]{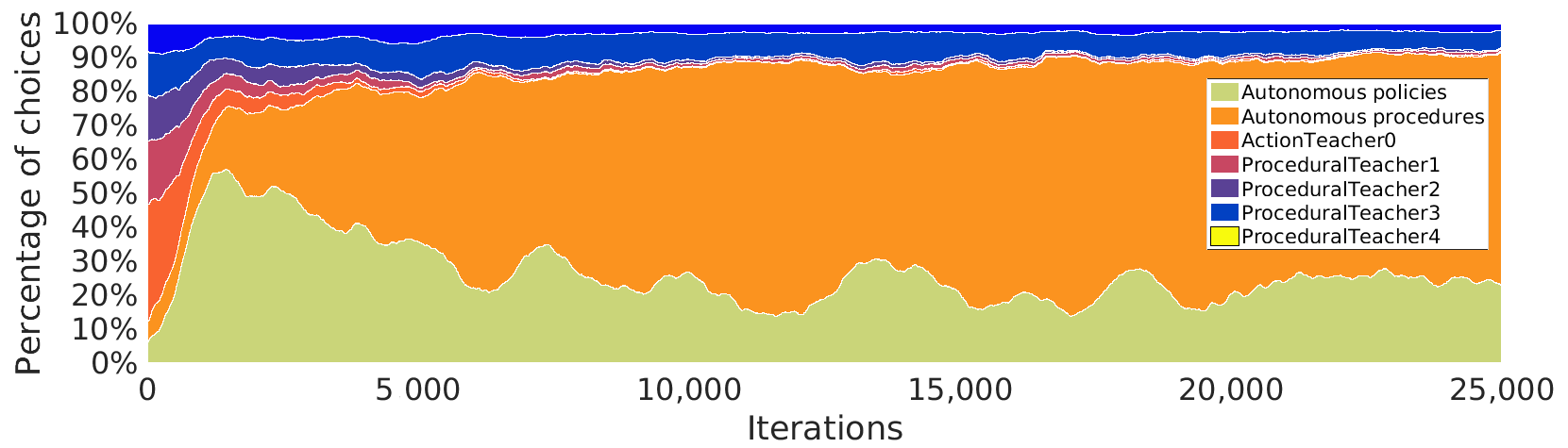}
\caption{{Evolution} 
 of choices of strategies for the SGIM-PB learner during the  learning process on the simulation setup.}
\label{fig:simu_yumi_strats}
\end{figure}

The performance per type of tasks in Figure~\ref{fig:physical_task_evaluation} shows that like in the simulation setup,  there is little difference for the simple tasks $\Omega_0$ and $\Omega_1$, and there is more difference on the complex tasks $\Omega_3$ and $\Omega_4$.   The more complex the task, the more SGIM-PB outperforms SGIM-ACTS.
These confirm that \textbf{procedural teachers are better adapted to tackle the most complex and hierarchical outcome spaces}. 

 \begin{figure}[H]
\includegraphics[width=0.95\linewidth]{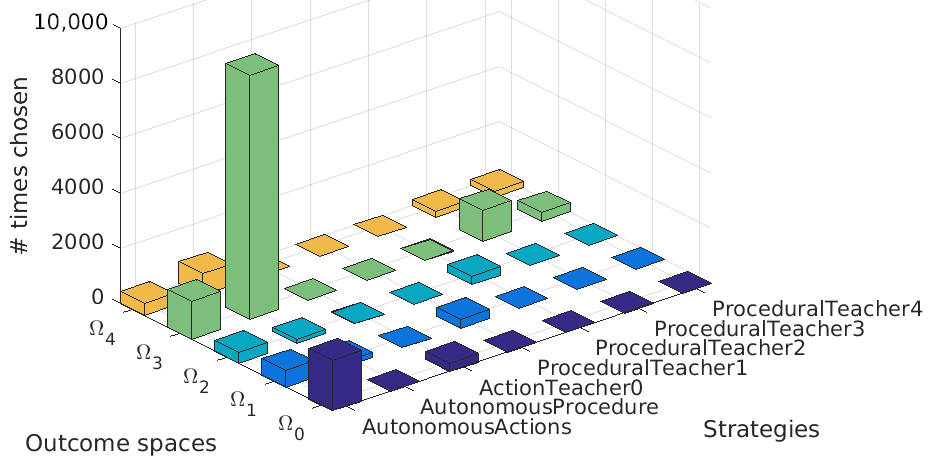}
\caption{Choices 
 of strategy and goal outcome for the SGIM-PB learner during the learning process on the simulation setup.}
\label{fig:simu_yumi_choices}
\end{figure}
\vspace{-6pt}
\begin{figure}[H]
\includegraphics[width=0.95\linewidth]{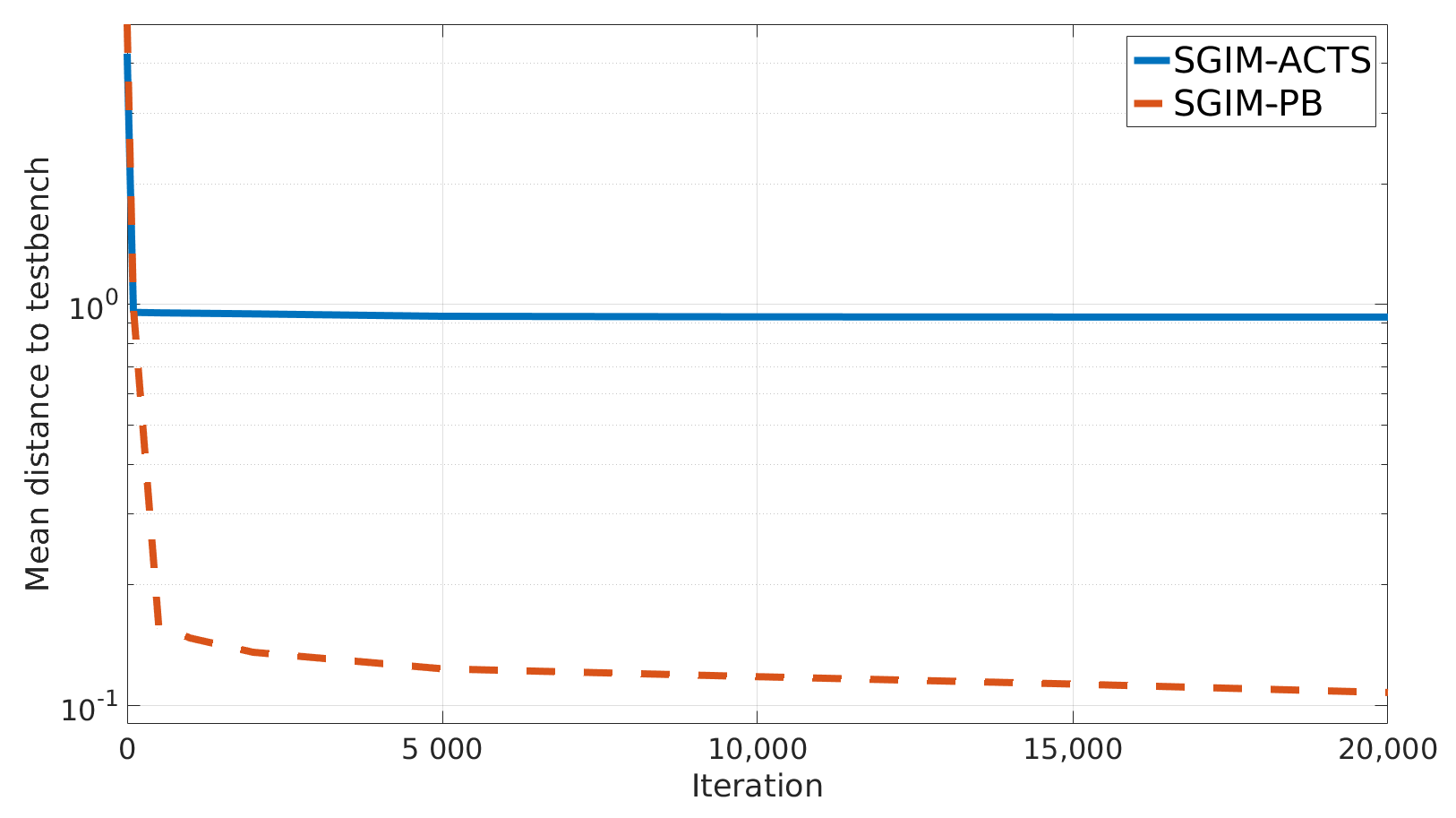}
\caption{{Evaluation} 
 of all algorithms throughout the learning process for the physical setup.}
\label{fig:real_yumi_evaluation}
\end{figure}
\vspace{-6pt}
\begin{figure}[H]
\includegraphics[width=0.95\linewidth]{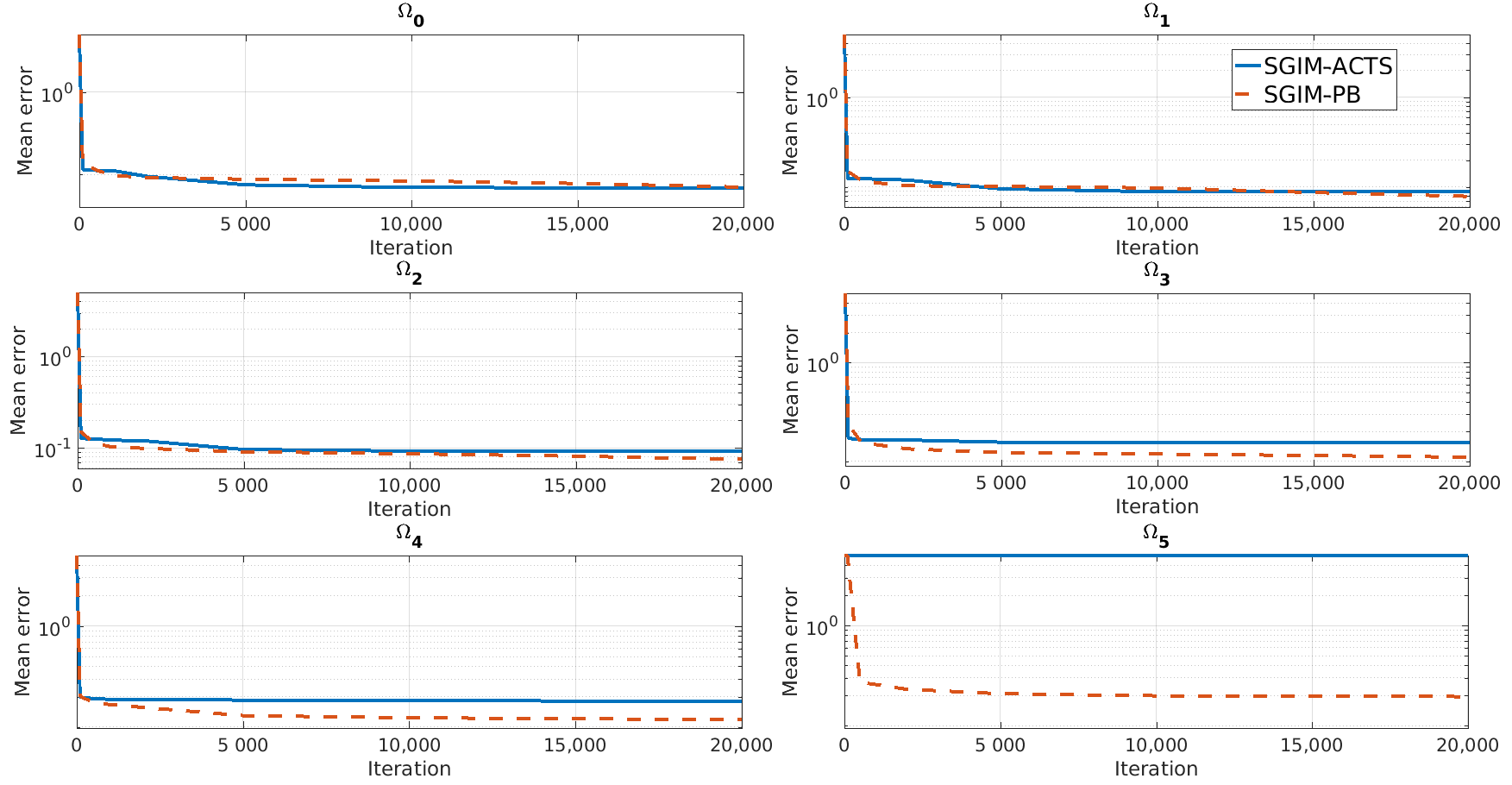}
\caption{{Evaluation} 
 {for} 
 each outcome space of the physical Yumi setup.}
\label{fig:physical_task_evaluation}
\end{figure}

\subsubsection{Autonomous Exploration of Procedures}
Moreover, in the real yumi setup, we have a supplementary type of tasks $\Omega_5$, which are also the highest level of hierarchy tasks.  For $\Omega_5$, no action or procedural teacher was provided to SGIM-PB or SGIM-ACTS, therefore we can contrast the specific effects of autonomous exploration of procedures to autonomous exploration of the action space.  Figure~\ref{fig:physical_task_evaluation} shows that the performance of SGIM-ACTS is constant for $\Omega_5$, it is not able to reach any task in $\Omega_5$ even once, while SGIM-PB, owing to the capability of the procedure framework to reuse the knowledge acquired for the other tasks, is able to explore in this outcome space.

To understand the reasons for this difference, let us examine the histogram of the strategies chosen per task in Figure~\ref{fig:physical_task_strategy}. For $\Omega_5$, SGIM-PB uses massively procedure space exploration compared to action space exploration. Owing to autonomous procedure exploration, SGIM-PB can thus learn complex tasks by using the decomposition of tasks into known subtasks.

\begin{figure}[H]
\includegraphics[width=0.9\linewidth]{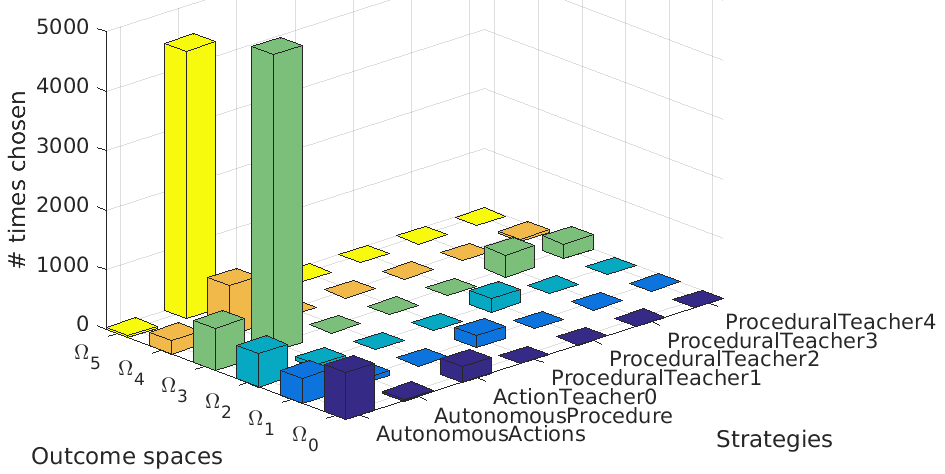}
\caption{Choices of strategy and goal outcome for the SGIM-PB learner during the learning process on the physical setup.}
\label{fig:physical_task_strategy}
\end{figure}


This highlights the \textbf{essential role of the procedure representation and the procedure space exploration by imitation but also by autonomous exploration, in order to learn high-level hierarchy tasks, which have sparse rewards}. 

\subsection{Curriculum Learning by SGIM-PB}
 Given the properties of imitation and procedures that we outlined, is SGIM-PB capable of choosing the right strategy for each task to build a curriculum starting from simple tasks before complex tasks?
 
Figure~\ref{fig:physical_task_strategy} shows that SGIM-PB uses more procedural exploration and imitation than action exploration or imitation for $\Omega_3$ and $\Omega_4$, compared to the more simple tasks. A coupling  appears between simple tasks and action space exploration on the one hand, and complex tasks and procedure exploration on the other hand. Moreover, for imitation, \textbf{SGIM-PB was overall able to correctly identify the most adapted teacher to each outcome space}. Their only suboptimal choice is to use the procedural teacher built for $\Omega_4$ to explore $\Omega_3$. This mistake can be explained as both outcome spaces have the same task decomposition (see Figure~\ref{fig:hierarchy}). 

Likewise, for the simulation setup, in Figure~\ref{fig:simu_yumi_choices}, the histogram of explored task-strategy combinations confirms that the learner uses mostly autonomous exploration strategies. 
It uses mostly action exploration  for the simplest outcome spaces ($\Omega_0, \Omega_1, \Omega_2$), and procedure exploration for the most complex outcomes ($\Omega_3, \Omega_4$). {This shows the complementarity of action and procedures for exploration in an environment with a hierarchical set of outcome spaces}. 
We can see for each task subspace, the proportion of imitation used. While for $\Omega_0$, SGIM-PB uses the strategy AutonomousActions five times more than ActionTeacher0, the proportion of imitation increases for the complex tasks. Imitation seems to be required more for complex tasks.
For the simplest outcome spaces ($\Omega_0, \Omega_1, \Omega_2$), it uses mostly action exploration  and procedure exploration for the most complex ones ($\Omega_3, \Omega_4$). 
From Figure~\ref{fig:simu_yumi_choices} , we can also confirm that for each task, the teacher most requested is specialised in the goal task, the only exceptions are ProceduralTeacher3 and ProceduralTeacher4 who are specialised in different complex tasks but that use the same subtask decomposition, thus demonstrations of ProceduralTeacher3 has effects on $\Omega_4$ and vice versa. The choices shown in the histogram show that SGIM-PB has spotted the teacher's domain of expertise. 

Let us analyse the evolution of the time each outcome space (Figure~\ref{fig:simu_yumi_tasks}) and strategy (Figure~\ref{fig:simu_yumi_strats}) is chosen  during the learning process of  SGIM-PB  on the simulated setup.  
In Figure~\ref{fig:simu_yumi_tasks}, its self-assigned curriculum starts by exploring the most simple task $\Omega_0$  until 1000 iterations. In Figure~\ref{fig:simu_yumi_strats}, we see that this period corresponds mainly to the use of the strategy Autonomous policies, relying on itself to acquire its body schema.  
Then it gradually switches to working on the most complex outcome space $\Omega_3$ (the highest dimension) and marginally more on $\Omega_4$ while preferring autonomous procedures and marginally the teachers for $\Omega_3$ and $\Omega_4$. In contrast, the strategy ActionTeacher0 decreases, SGIM-PB does not use action imitation any more. \textbf{SGIM-PB switches from imitation of action primitives to procedures, and most of all it turns to the strategies autonomous policies and autonomous procedures}.

For the physical setup, the evolution of outcome spaces (Figure~\ref{fig:physical_tasks}) and strategies (Figure~\ref{fig:physical_strats}) chosen are more difficult to analyse. However, they show the same trend from iterations 0 to 10,000: the easy outcome spaces  $\Omega_0, \Omega_1, \Omega_2$ are more explored in the beginning before being neglected after 1000 iterations to explore the most complex outcome spaces $\Omega_3, \Omega_5$.  Imitation is mostly used in the beginning of the learning process}, whereas later in the curriculum, autonomous exploration is preferred. The autonomous exploration of actions strategy was also less used. However, the curriculum after 10,000 is harder to analyse : at two instances (around 13,000 and a second time around 17,000 iterations), SGIM-PB switched to autonomous exploration of actions while  exploring simpler outcome spaces $\Omega_0, \Omega_1, \Omega_2$. This might mean the learner needs to consolidate its basic knowledge on the basic tasks before being able to make further progress in the complex tasks.

\begin{figure}[H]
\includegraphics[width=0.9\linewidth]{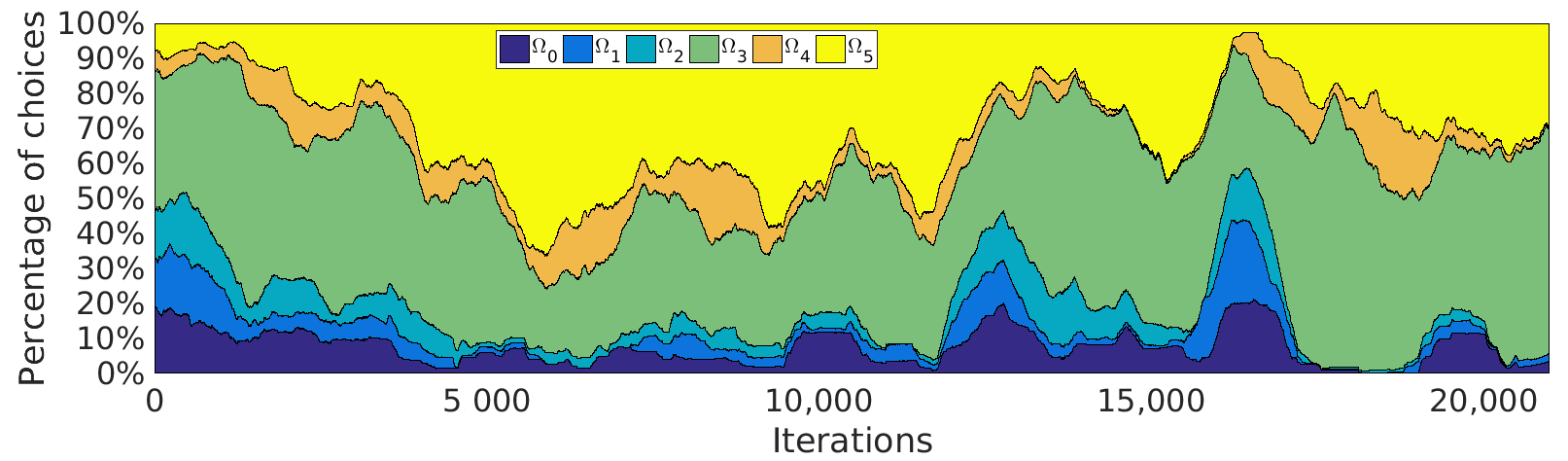}
\caption{{Evolution} 
 of choices of tasks for the SGIM-PB learner during the  learning process on the physical setup.}
\label{fig:physical_tasks}
\end{figure}
\vspace{-6pt}
\begin{figure}[H]
\includegraphics[width=0.9\linewidth]{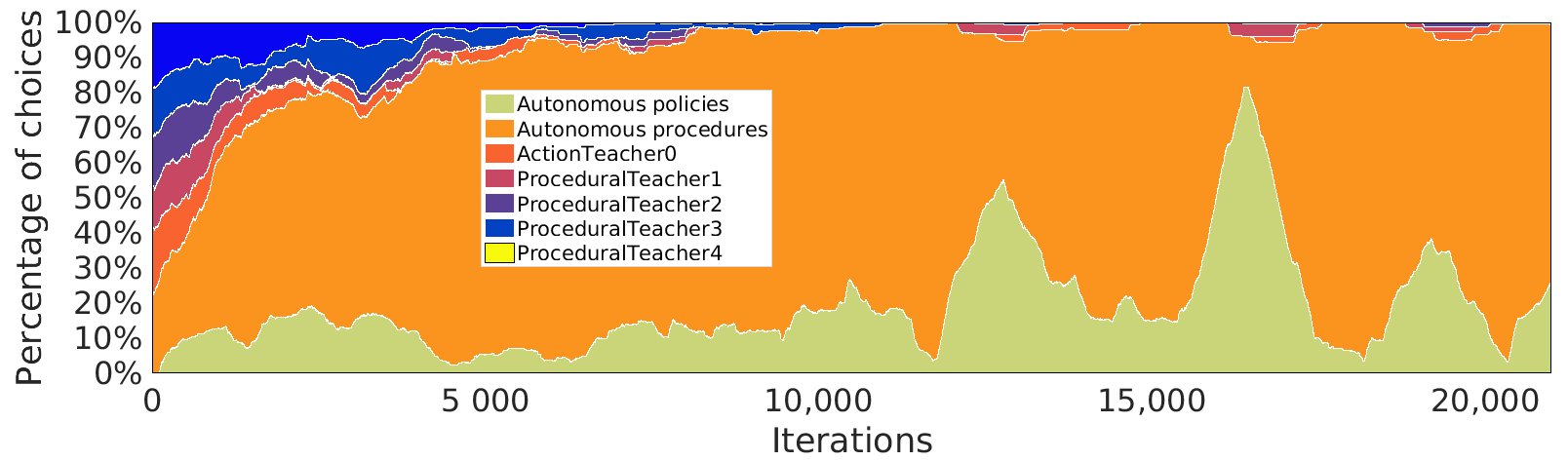}
\caption{{Evolution} 
 of choices of strategies for the SGIM-PB learner during the  learning process on the physical setup.}
\label{fig:physical_strats}
\end{figure}

\subsection{Data Efficiency of Active Learning:  Active Imitation Learning vs Batch Transfer}

\label{sec:tlSetup}

In this section, we explore the possibility to transfer a set of procedures as a batch at the beginning of the learning process, as opposed to active requests from the learner throughout the learning process. We consider a learner and a teacher with different embodiments working on the same tasks.  While transfer of knowledge of actions can not be reused straightforward, how can the knowledge of procedures be exploited? 

  In our example, we consider new learners trying to explore the interactive table using the left arm of Yumi, while benefitting from the knowledge acquired from a right arm Yumi. We call this simulation setup the \textit{left-arm setup}. We extracted a dataset $D_3$ composed of the procedures and their corresponding reached outcomes $((\omega_i, \omega_j), \omega_r))$ taken from the experience memory of a SGIM-PB learner that has trained on the right arm for 25,000 iterations (taken from the runs of SGIM-PB on the simulated setup). To analyse the benefits of batch transfer before learning, we run in the simulated setup (see Section~\ref{sec:setup}), two variants of SGIM-PB 10 times, with for each 10,000 iterations:

\begin{itemize}
\item Left-Yumi-SGIM-PB: the classical SGIM-PB learner using its left arm, using from the exact same strategies as on the simulated setup, without any procedure transferred;
\item Left-Yumi-SGIM-TL: a modified SGIM-PB learner, benefiting from the strategies used on the simulated setup, and which benefits from the dataset $D_3$ as a Transferred Lump at the initialisation phase:  $Memo \gets \{ ((\omega_i, \omega_j), \omega_r) \}$ at the beginning their learning process. \resubmit{No actions are transferred, and the transferred data are only used for computing local exploration of the procedural space, so they don't impact the interest model nor the test evaluations reported in the next section.} 
\end{itemize}

All procedural teachers propose the same procedures as in Section~\ref{sec:setup} for the right-handed SGIM-PB learner. The ActionTeacher0, proposing action demonstrations for reaching $\Omega_0$ outcomes was changed to 
left-handed demonstrations.

\subsubsection{Evaluation Performance}

Figure~\ref{fig:transfer_evaluation} shows the global evaluation of Left-Yumi-SGIM-PB and Left-Yumi-SGIM-TL. We can see that, even though the learning process seems quicker before 500 iterations for the Left-Yumi-SGIM-TL learner, both learners quickly learn at the same rate as shown by their overlapping evaluation graphs. If we look at Figure~\ref{fig:transfer_evaluation_tasks} which shows this evaluation broken down for each task, we see the same phenomenon for  $\Omega_2,\Omega_3$ and $\Omega_4$. This seems to indicate, that active imitation using the small demonstration datasets are enough for Left-Yumi-SGIM-PB to tackle this setup, while the huge dataset of transferred procedures don't give Left-Yumi-SGIM-TL an edge other than a slightly better performance for the complex tasks at the initial phase.

\begin{figure}[H]
\includegraphics[width=0.9\linewidth]{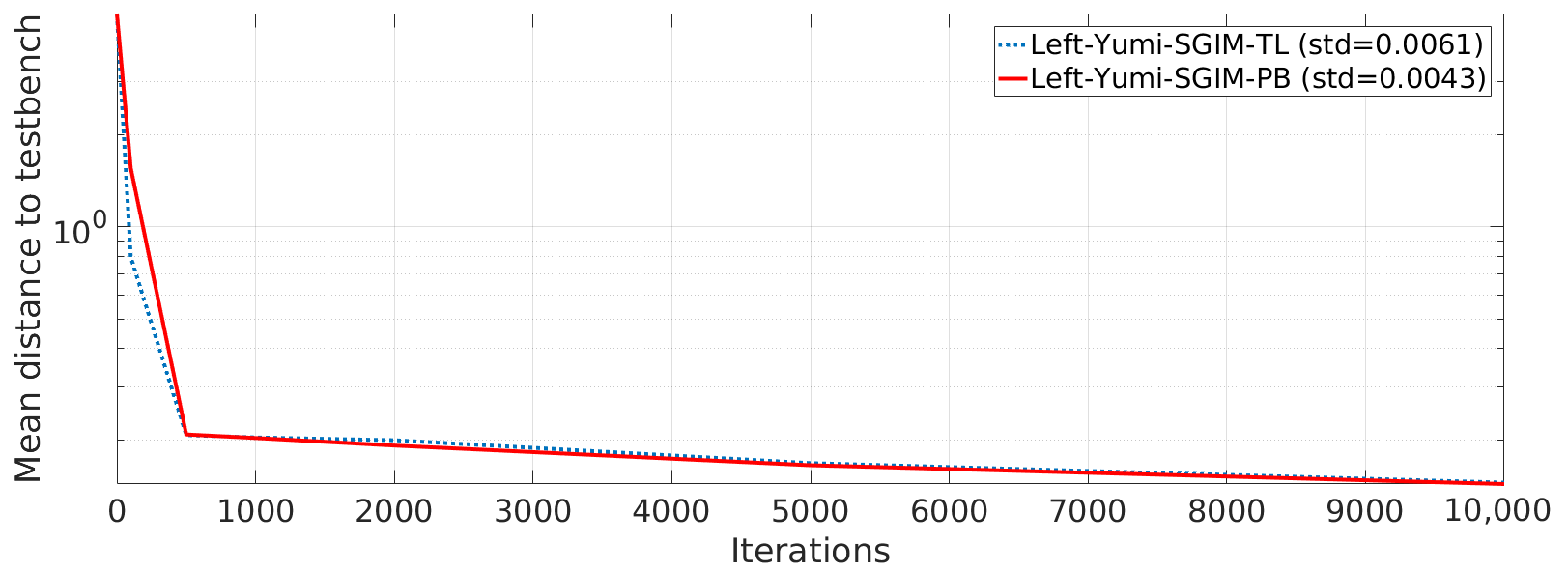}
\caption{\resubmit{Evaluation 
 of all algorithms throughout learning process for the transfer learning setup, final standard deviations are given in the legend}.}
\label{fig:transfer_evaluation}
\end{figure}

\subsubsection{Procedures Learned}

To analyse what task hierarchy both learners have discovered at the end of their learning process, we plotted Figure~\ref{fig:transfer_proc_task}. We can see that no learner is clearly better at discovering the setup task hierarchy (see Figure~\ref{fig:hierarchy}): Left-Yumi-SGIM-PB decomposes more often tasks $\Omega_1$ as $(\Omega_0,\Omega_0)$, but slightly less often  $\Omega_3$ or $\Omega_4$ as $(\Omega_1,\Omega_2)$. For $\Omega_3$, Left-Yumi-SGIM-PB also uses another procedure: $(\Omega_2,\Omega_1)$ which is also valid. To position both objects, the robot can start by either object 1 or object 2. However, if we take into account the task hierarchy learned by the transfer dataset which is fed to Left-Yumi-SGIM-TL before its learning process, we can see that Left-Yumi-SGIM-TL has learned almost exactly the same task hierarchy than the transfer dataset, which indicates its influence owing to the large number of procedures transferred: procedures from 25,000 iterations were transferred compared to the new procedures explored by Left-Yumi-SGIM-TL during only 10,000 iterations. Hence Left-Yumi-SGIM-TL has the same defects and qualities than the transfer dataset in terms of the task hierarchy discovered. For instance, Left-Yumi-SGIM-TL uses the inefficient procedure $(\Omega_1,\Omega_2)$ to reach $\Omega_1$ more often than Left-Yumi-SGIM-PB. It wasn't able to overstep the defects of the transfer dataset.

\begin{figure}[H]
\includegraphics[width=0.9\linewidth]{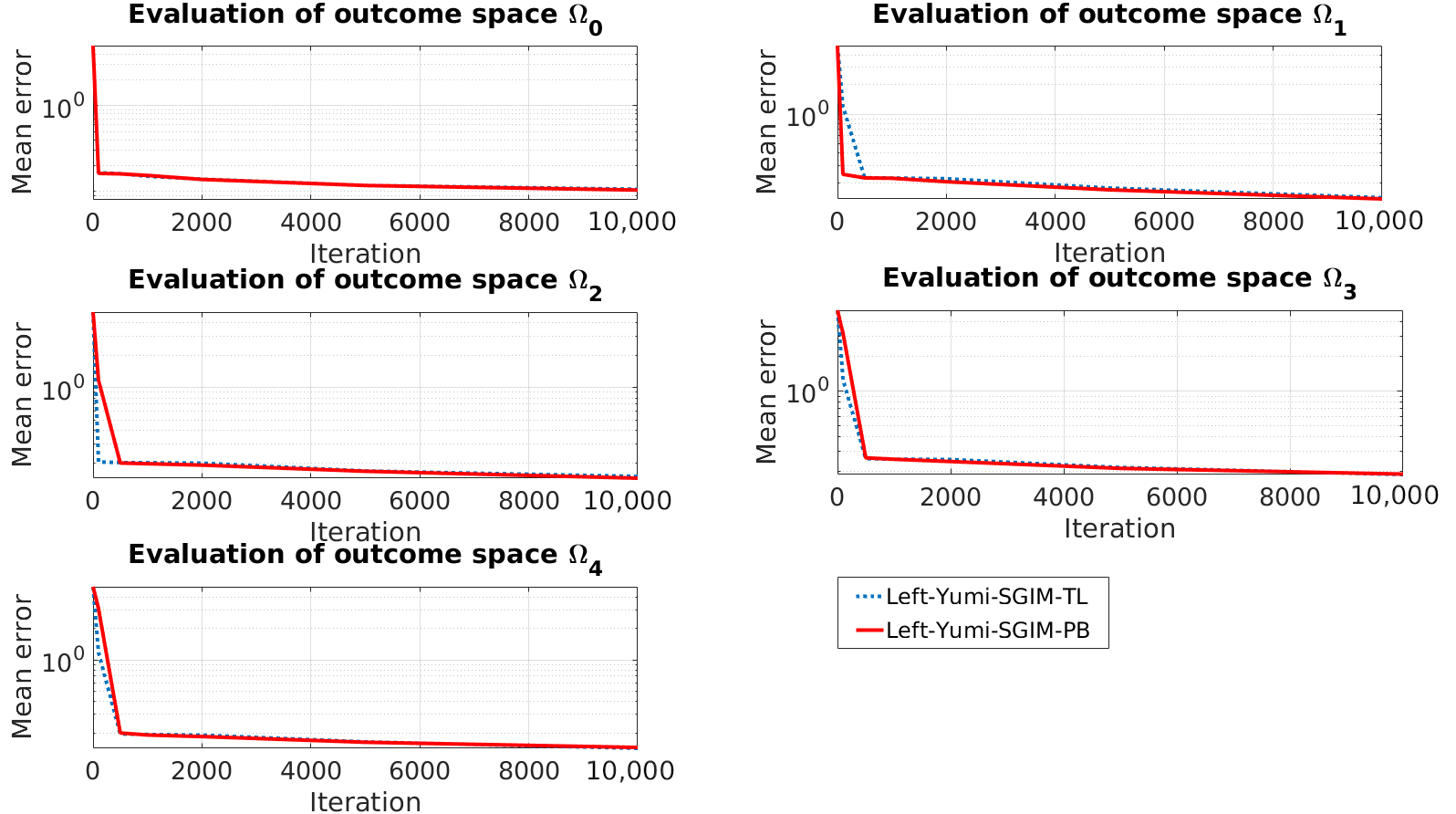}
\caption{\resubmit{Task 
 evaluation of all algorithms throughout learning process for transfer learning setup}.}
\label{fig:transfer_evaluation_tasks}
\end{figure}

\subsubsection{Procedures Used during Learning}

If we look at the procedures that were actually tried out by the learner during its learning process for reaching each goal outcome (see Figure~\ref{fig:transfer_proc_used}), we can see first that {for all versions of the algorithm and all types of tasks except $\Omega_1$, the procedures most used during the learning phase correspond to the ground truth. Thus intrinsic motivation has oriented the exploration towards the relevant task decompositions.} 
Besides, for all types of task, Left-Yumi-SGIM-TL tends to explore more the procedures in $(\Omega_1, \Omega_2)$ than Left-Yumi-SGIM-PB. This difference can also be explained by the predominance of this procedure in the transfer dataset $D_3$. Both of them also tried a lot of procedures from the $\Omega_0^2$ procedural space. It confirms that Left-Yumi-SGIM-TL was not able to filter out the defects of its transfer dataset in terms of the task hierarchy provided.

\subsubsection{Strategical Choices}

Analysing the learning process (Figures~\ref{fig:transfer_SGIM_PB_strats}--\ref{fig:transfer_SGIM_TL_tasks}) of both learners, we can see that they are very similar. Both learners start by performing a lot of imitation of the available teachers coupled with an exploration of all outcome types, until 1500 and 2000 iterations for respectively Left-Yumi-SGIM-TL and Left-Yumi-SGIM-PB. This difference in timing can be caused by the transferred dataset $D_3$, but is not very significant. Then they focus more on the autonomous exploration of the action space to reach the $\Omega_0$ outcome subspace before gradually working on more complex outcome spaces while performing more an more autonomous exploration of the procedural space. However, the Left-Yumi-SGIM-TL learner seems to abandon its initial imitation phase faster than Left-Yumi-SGIM-PB (about 1000~iterations faster), and also quickly starts working on the more complex $\Omega_3$ outcome space with the strategy autonomous exploration on procedures. This initial faster maturation seems perhaps too fast as Left-Yumi-SGIM-TL reverts to working on $\Omega_0$ with autonomous actions afterwards : we see two other peaks of this choice of combination at 5500 iterations and 9500 iterations. On the contrary,  Left-Yumi-SGIM-PB  seems to converge more steadily towards its final learning phase.

\begin{figure}[H]
\includegraphics[width=0.98\linewidth]{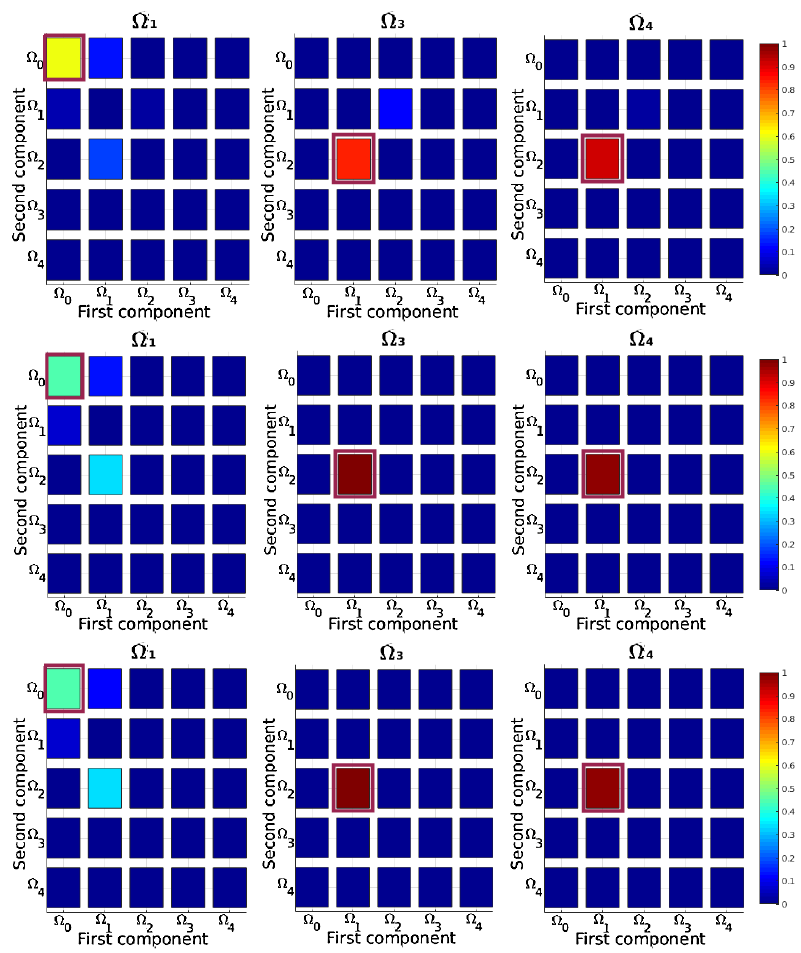}
\caption{\resubmit{Task hierarchy discovered by the learners on left-arm setup: this represents for outcome spaces $\Omega_1,\Omega_3$ and $\Omega_4$ the percentage of time each procedural space is chosen during test to reach goals of the testbench by  Left-Yumi-SGIM-PB  (\textbf{1st row}), the transfer dataset (\textbf{2nd row}) and the Left-Yumi-SGIM-TL learner (\textbf{3rd row})}. {The ground truth hierarchy is indicated by a red square.}}
\label{fig:transfer_proc_task}
\end{figure}
\vspace{-6pt}

\begin{figure}[H]
\includegraphics[width=0.95\linewidth]{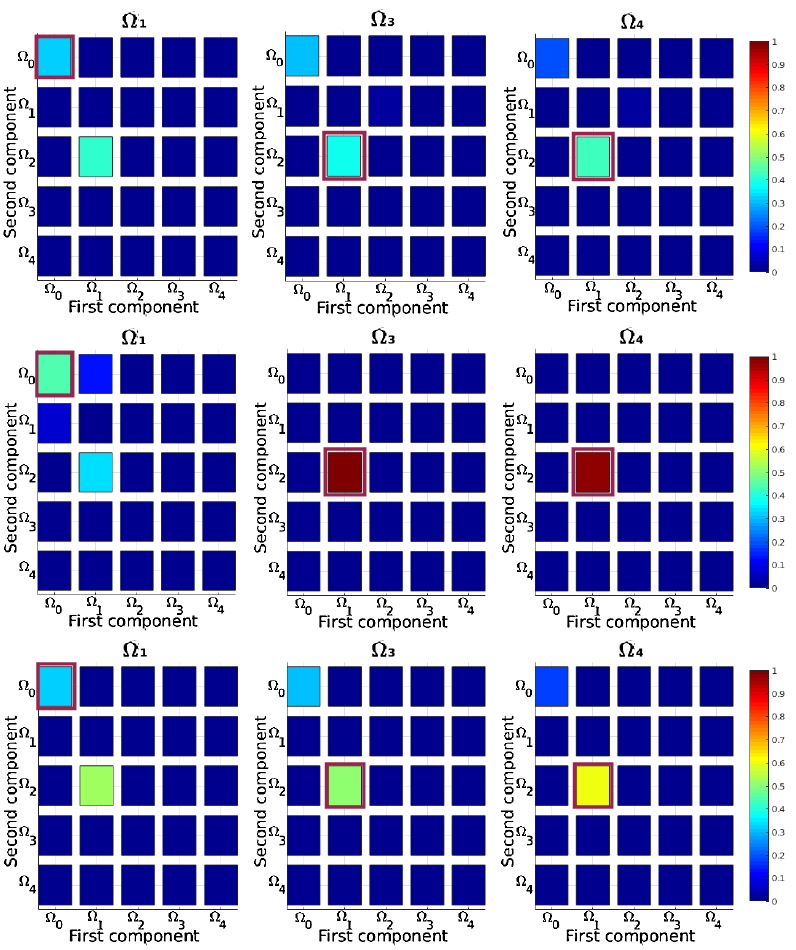}
\caption{\resubmit{Procedures used during the exploration phase on left-arm setup: this shows for outcome spaces $\Omega_1, \Omega_3$ and $\Omega_4$ the percentage of times each procedural space was chosen during learning by Left-Yumi-SGIM-PB (\textbf{1st row}), and the Left-Yumi-SGIM-TL learner (\textbf{3rd row}), procedures discovered by the transfer dataset $\mathcal{D}_3$ are also reminded (\textbf{2nd row})}. {The ground truth hierarchy is indicated by a red square.}}
\label{fig:transfer_proc_used}
\end{figure}

\begin{figure}[H]
\includegraphics[width=0.95\linewidth]{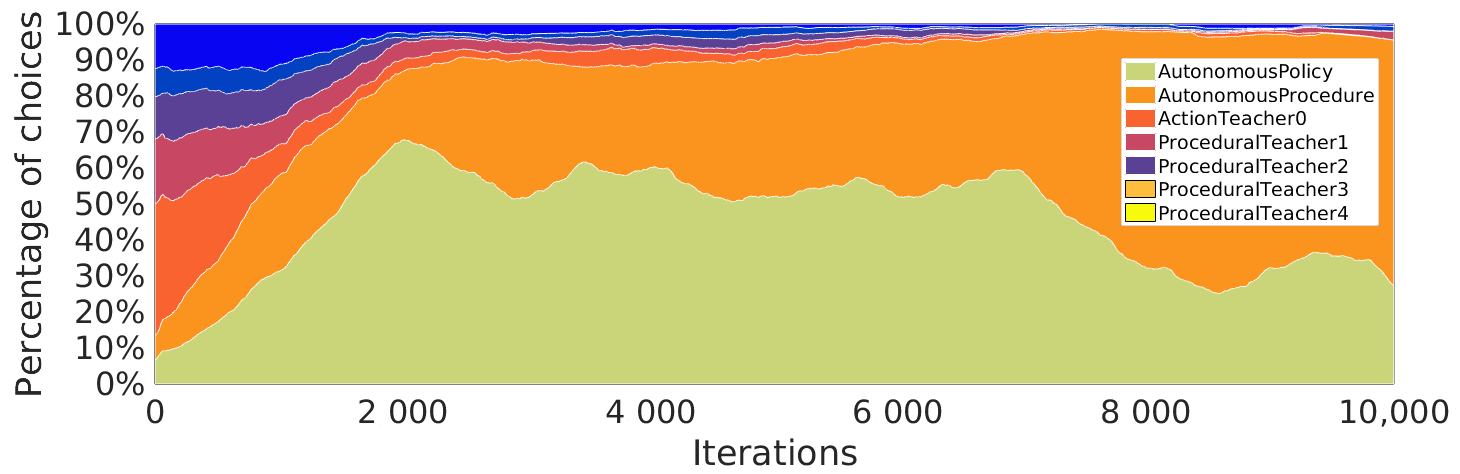}
\caption{\resubmit{Evolution 
 of choices of strategies for the Left-Yumi-SGIM-PB learner during the learning process on left-arm setup}.}
\label{fig:transfer_SGIM_PB_strats}
\end{figure}

\begin{figure}[H]
\includegraphics[width=\linewidth]{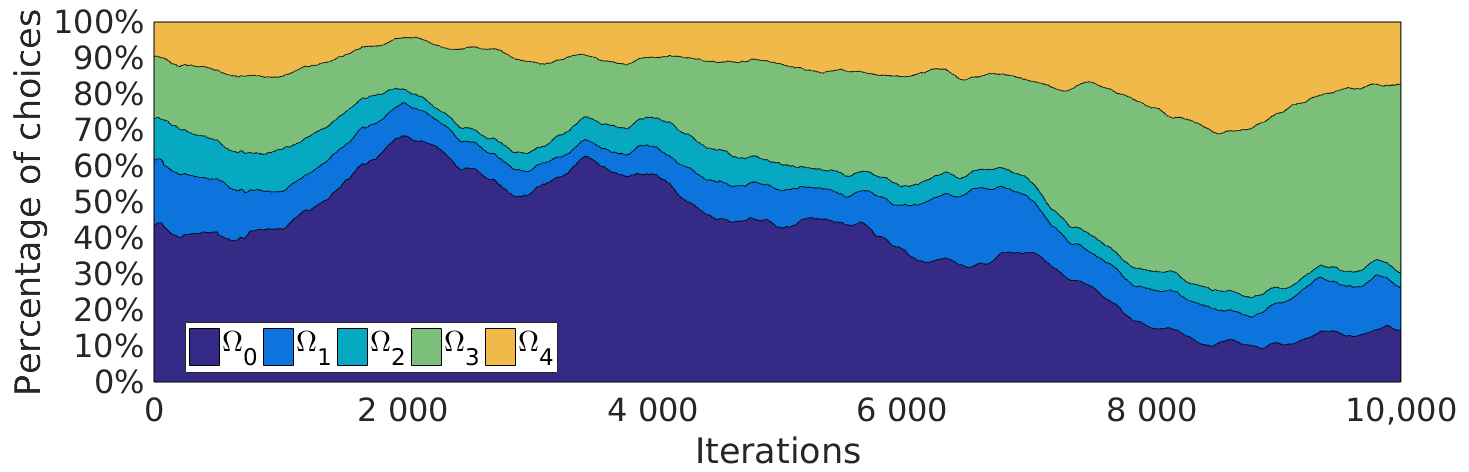}
\caption{{Evolution 
 of choices of tasks for the Left-Yumi-SGIM-PB learner during the learning process on left-arm setup}.}
\label{fig:transfer_SGIM_PB_tasks}
\end{figure}

\begin{figure}[H]
\includegraphics[width=\linewidth]{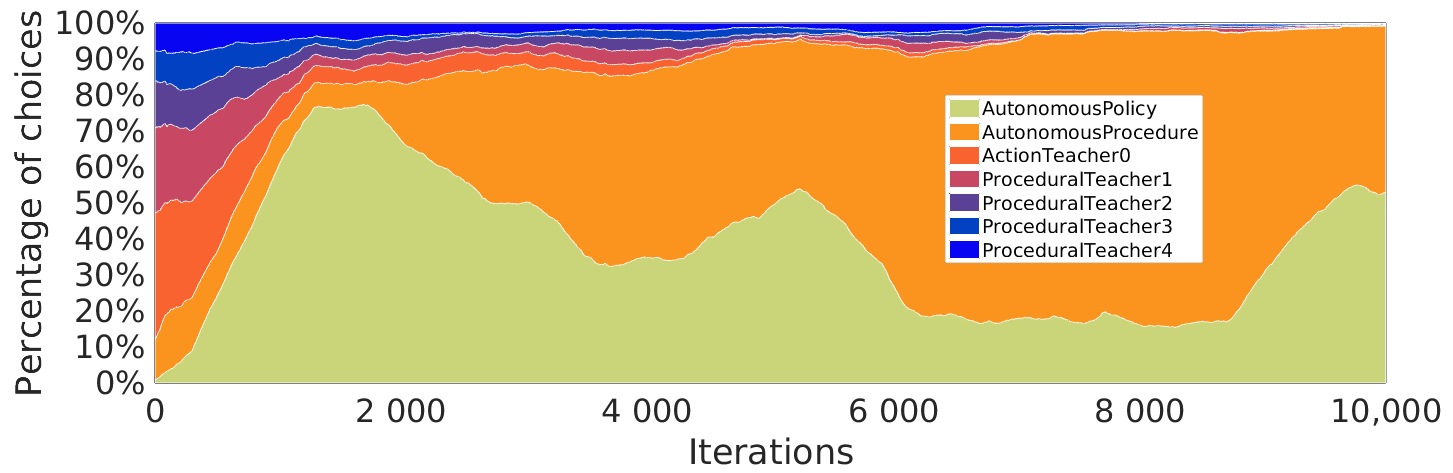}
\caption{\resubmit{Evolution 
 of choices of strategies for the Left-Yumi-SGIM-TL learner during the learning process on left-arm setup}.}
\label{fig:transfer_SGIM_TL_strats}
\end{figure}

\begin{figure}[H]
\includegraphics[width=\linewidth]{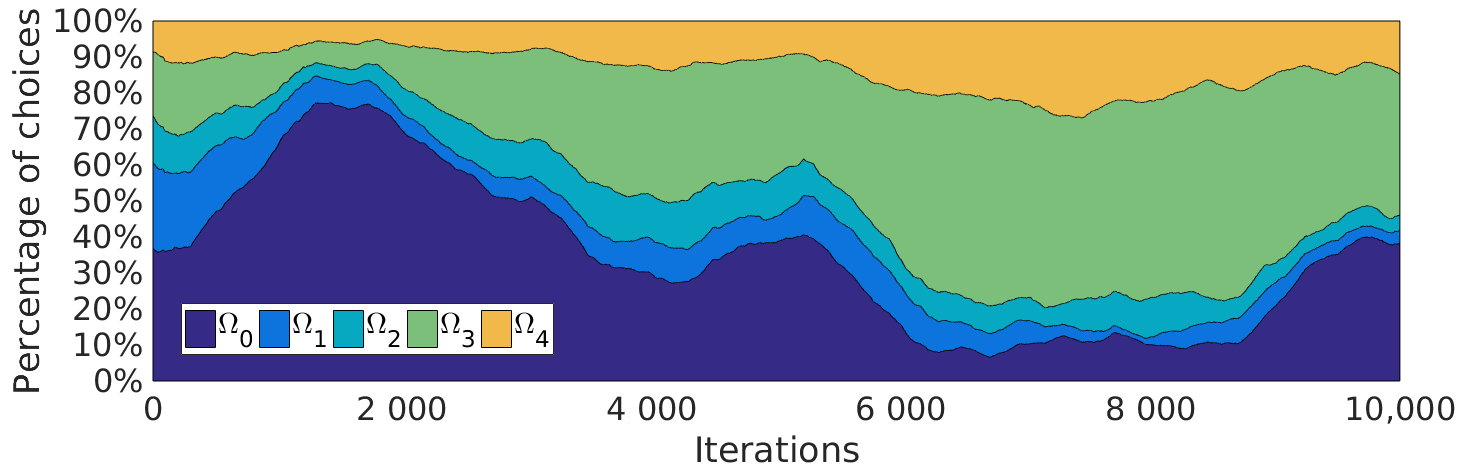}
\caption{{Evolution 
 of choices of tasks for the Left-Yumi-SGIM-TL learner during the learning process on left-arm setup}.}
\label{fig:transfer_SGIM_TL_tasks}
\end{figure}

If we look at the choices of strategy and goal outcomes for the whole learning process (see Figures~\ref{fig:transfer_SGIM_PB_task_strats} and \ref{fig:transfer_SGIM_TL_task_strats}), we can see that this difference in the learning processes is visible in the number of times each of the two main combinations of task and goal outcome space was chosen: Left-Yumi-SGIM-TL favors more working autonomously on procedures for exploring $\Omega_3$ whereas Left-Yumi-SGIM-PB worked more on autonomous exploration of actions for $\Omega_0$.

\begin{figure}[H]
\includegraphics[width=0.95\linewidth]{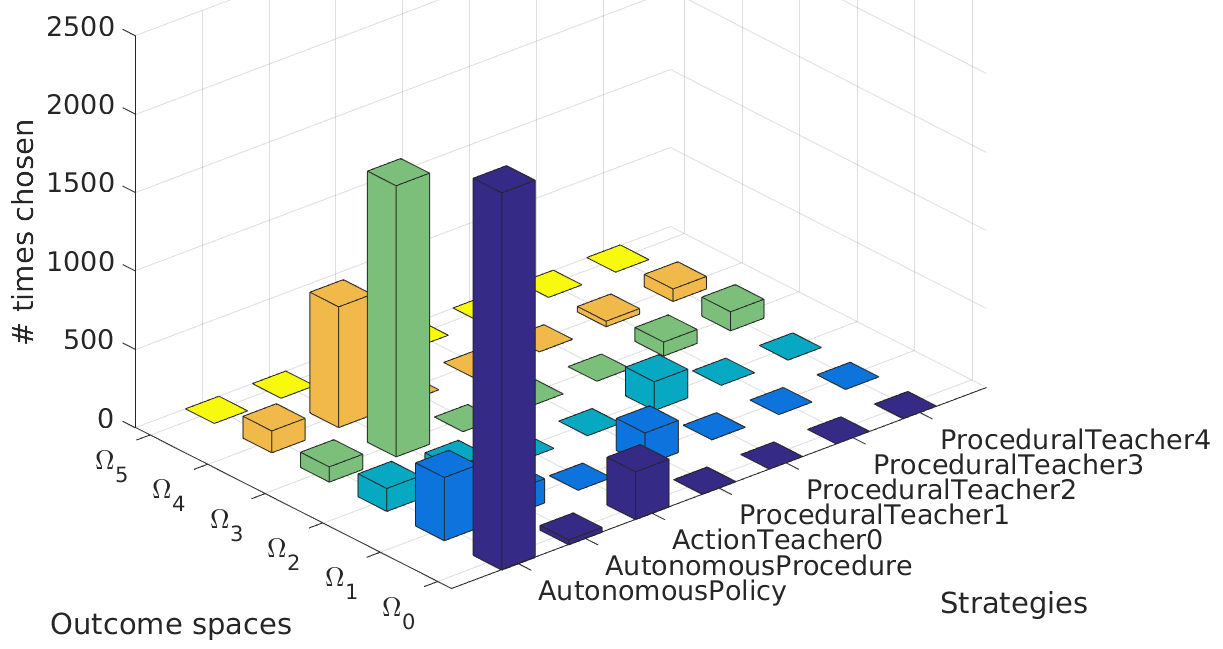}
\caption{{\textls[-20]{Choices of strategy and goal outcome for the Left-Yumi-SGIM-PB learner on left-arm setup}}.}
\label{fig:transfer_SGIM_PB_task_strats}
\end{figure}
\vspace{-6pt}
\begin{figure}[H]
\includegraphics[width=0.95\linewidth]{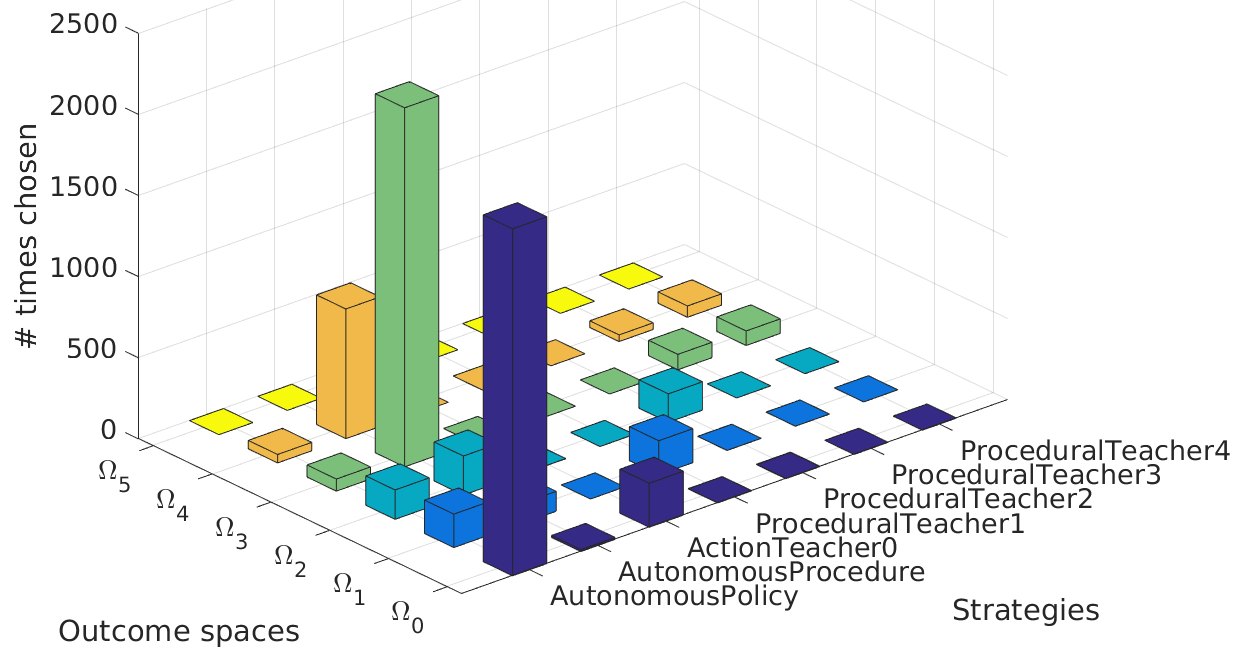}
\caption{\textls[-20]{Choices of strategy and goal outcome for the Left-Yumi-SGIM-TL learner on the left-arm~setup}.}
\label{fig:transfer_SGIM_TL_task_strats}
\end{figure}

These results seem to indicate that if a transfer of knowledge about difficult tasks takes place before easy tasks are learned, it can disturb the learning curriculum by changing the learner's focus on difficult tasks. The learner needs to realise that the gap in knowledge is too high, and give up these difficult tasks, to re-focus on learning the easy subtasks. 
Thus, demonstrations given throughout the learning process and adapted to the development of the learner seem more beneficial than a batch of data given at a single point of time, despite a larger amount of data. These results show that \textbf{procedure demonstrations can be effective for robots of different embodiments to learn complex tasks; and active requests of few procedure demonstrations are effective to learn task hierarchy}. They were not significantly improved by more data added at initialisation in terms of error, exploration strategy or autonomous curriculum learning. 

\section{Discussion} 
\label{sec:discussion}


The experimental results highlight the following properties of SGIM-PB:
\begin{itemize}
\item procedures representation and task composition become necessary to reach tasks of higher hierarchy. It is not a simple bootstrapping effect.
\item transfer of knowledge of procedures is efficient for cross-learner transfer of knowledge even when they have different embodiments.
\item active imitation learning of procedures is more advantageous than imitation from a dataset provided from the initialization phase.
\end{itemize}

The  performance of SGIM-PB stems from its tackling several aspects of transfer of knowledge, and relies on our proposed representation of compound actions that allows hierarchical reinforcement learning.

\subsection{A Goal-Oriented Representation of Actions}
Our representation of actions allows our learning algorithm to \textbf{adapt the complexity of its actions to the complexity of the task at hand}, whereas other approaches using via points~\cite{Stulp2011H}
, or parametrised skills~\cite{Silva20122ICMLI2}, had to bound the complexity of their actions. 
Other approaches like~\cite{options} 
 also use a temporally abstract representation of actions:  options. 
However options are often used in discrete states settings to reach 
 specific states such as bottlenecks, and are generally learned beforehand. On the contrary, our dual representation of skills as action primitive sequences and procedures allow an online learning of an unlimited number of complex behaviours. Our exploitation of the learned action primitive sequence is simplistic and needs improvement though.

Our work proposes a representation of the relationship between tasks and their complexities, by the procedural framework. 
{Comparatively,}~\cite{Manoury2019HAI} proposed for tackling a hierarchical multi-task setting, to learn action primitives and use planning to recursively chain skills. 
However, {that} approach does not build a representation of a sequence of action primitives, and  planning grows slower as the environment is explored. A conjoint use of planning techniques with an unrestrained exploration of the procedural space could be an interesting prospect and extension of our work.

In this article, we tackled the learning of complex control tasks using sequences of actions of unbounded length. Our main contribution is a \textbf{dual representation of compound actions in both action and outcome spaces}. We showed its impact on autonomous exploration but also on imitation learning: SGIM-PB learns the most complex tasks by autonomous procedural space exploration, and can benefit more from procedural demonstrations for complex tasks and from motor demonstrations for simple tasks{, confirming the results on a more simple setup in~\cite{Duminy2019FN}}. Our work demonstrates the gains that can be achieved by requesting just a small amount of demonstration data with the right type of information with respect to the complexity of the tasks. Our work should be improved by a better exploitation algorithm of the low-level control model and can be speeded up by adding planning.

\subsection{Transfers of Knowledge}

We have tested our algorithm in the classical setups of transfer of knowledge :  \textbf{cross-task transfer, cross-learner transfer and by imitation learning}. We have shown that \textbf{SGIM-PB can autonomously determine} the main questions of Transfer Learning as theorised in~\cite{Pan2010ITKDE}:
\begin{itemize}
\item What information to transfer? For compositional tasks, a demonstration of task decomposition is more useful than {a demonstration} of an action, as it helps bootstrap cross-task transfer of knowledge. Our {case study shows} a clear advantage of procedure demonstrations and {procedure exploration}. 
 On the contrary, for simple tasks, action demonstrations and action space exploration show more advantages. Furthermore, for cross-learner transfer, especially when the learners have different embodiments, {this case study indicates} that demonstrations of procedures are still helpful{,} whereas demonstrations of actions {are no longer relevant}.
\item How to transfer? We showed that decomposition of a hierarchical task, through procedure exploration and imitation, is more efficient than {learning directly action parameters, i.e., interpolation of action parameters. This confirms the results found in a more simple setup in~\cite{Duminy2019FN}}.
\item When to transfer? Our last setup shows that 
 an active imitation learner asking adapted demonstrations, as its competence increases, {performs almost better than when it is given a significantly larger batch of data at initialisation time}. More generally {for a less data-hungry transfer learning algorithm}, the transferred {dataset should be given to the learner in a timely manner so that the information is} adapted to the current level of the learner, i.e., its zone of proximal development~\cite{Vygotsky1978}. {This advantage has already been shown by an active and strategic learner---SGIM-IM~\cite{Nguyen2012IICHR}, a simpler versions of SGIM-PB---which had better performance than passive learners  for multi-task learning using action primitives.}
\item Which source of information? Our strategical learner {chooses} for each task the most efficient strategy between self-exploration and imitation{. Most} of all, it could understand the domain of expertise of the different teachers and choose the most appropriate expert to imitate. {The results of this case study confirms the results found in~\cite{Duminy2019FN} in a simpler setup and in~\cite{Nguyen2012PJBR} for learning action primitives}. 
\end{itemize}

\section{Conclusions\label{sec:conclusion}}



We showed that our industrial robot could learn sequences of motor actions of unrestrained size to achieve a field of hierarchically organized outcomes. To learn to control in continuous high-dimensional spaces of outcomes through a continuous infinite dimensionality space of actions, we combined: goal-oriented exploration to enable the learner to organize its learning process in a multi-task learning setting, procedures as a task-oriented representation to build increasingly more complex sequences of actions, active imitation strategies to select the most appropriate information and source of information,  and intrinsic motivation as a heuristic to drive the robot's curriculum learning process.  All four aspects are combined inside a curriculum learner called SGIM-PB. This algorithm showed the following characteristics through this study:

\begin{itemize}
	\item Hierarchical RL: it learns online task decomposition on 4 levels of hierarchy using the procedural framework; and it exploits the task decomposition to match the complexity of the sequences of action primitives to the task;
	\item Curriculum learning: it autonomously switches from simple to complex tasks, and from exploration of actions for simple tasks to exploration of procedures for the most complex tasks;
	\item Imitation learning: it empirically infers which kind of information is useful for each kind of task and requests just a small amount of demonstrations with the right type of information by choosing between procedural and action teachers;
	\item Transfer of knowledge: it automatically decides what information, how, when to transfer and which source of information for cross-task and cross-learner transfer learning.
\end{itemize}

  Thus, \textbf{our work proposes an active imitation learning algorithm based on intrinsic motivation that uses empirical measures of competence progress to choose at the same time what target task to focus on, which source tasks to reuse and how to transfer knowledge about task decomposition.} 
 Our contributions, grounded in the field of cognitive robotics, are : 
 a new  representation of complex actions enabling the exploitation of task decomposition and the proposition for tutors of 
 supplying information on the task hierarchy to learn compound tasks. 

This work should be improved by a better exploitation algorithm of the low-level control model and can be speeded up by adding planning methods.



\vspace{6pt} 



\authorcontributions{Conceptualization, S.M.N.; Formal analysis, J.Z.; Funding acquisition, D.D.; Investigation, N.D.; Methodology, Jerome Kerdreux; Writing---original draft, N.D.; Writing---review \& editing, S.M.N. All authors have read and agreed to the published version of the manuscript.}

\funding{This work was supported by Minist\`ere de l'Education Nationale, de l'Enseignement Sup\'erieur et de la Recherche, European Regional Development Fund (ERDF), R\'egion Bretagne, Conseil G\'en\'eral du Finist\`ere) and by Institut Mines T\'el\'ecom, received in the framework of the VITAAL project.}

\institutionalreview{Not applicable
}

\informedconsent{Not applicable

}

\dataavailability{The codes used are available at \url{https://bitbucket.org/smartan117/sgim-yumi-simu} (simulated version), and at \url{https://bitbucket.org/smartan117/sgim-yumi-real} (physical~one).
} 

\acknowledgments{\textls[-20]{This work was supported by Minist\`ere de l'Education Nationale, de l'Enseignement Sup\'erieur et de la Recherche, European Regional Development Fund (ERDF), R\'egion Bretagne, Conseil G\'en\'eral du Finist\`ere) and by Institut Mines T\'el\'ecom, received in the framework of the VITAAL~project.}}

\conflictsofinterest{The authors declare no conflict of interest. The funders had no role in the design of the study; in the collection, analyses, or interpretation of data; in the writing of the manuscript, or in the decision to publish the results.} 


%
%
%
%
\end{paracol}
\reftitle{References}


\externalbibliography{yes}

%
%
%



\end{document}